\documentclass[12pt, letterpaper]{article}
\usepackage{verbatim}

\usepackage{PRIMEarxiv}

\usepackage[utf8]{inputenc}     
\usepackage[T1]{fontenc}        
\usepackage{multicol}

\usepackage{microtype}          
\usepackage{setspace}           
\usepackage{textcomp}           
\usepackage{lipsum}             

\usepackage{amsmath, amssymb, amsfonts, mathtools} 
\usepackage{eucal}              
\usepackage[version=4]{mhchem}             
\usepackage{nicefrac}           
\usepackage{nccmath}
\usepackage{bm}

\usepackage{graphicx}           
\graphicspath{{media/}}         
\usepackage{caption, subcaption}
\usepackage{float}              
\usepackage{adjustbox}          
\usepackage{booktabs}           
\usepackage{multirow}           
\usepackage{longtable}          
\usepackage{rotating}           
\usepackage{pdflscape}          
\usepackage{placeins}           

\usepackage{algorithm}          
\usepackage{algpseudocode}      

\usepackage{enumitem}           

\usepackage{url}                
\usepackage{hyperref}           

\usepackage{color,soul}         

\usepackage{fancyhdr}           
\usepackage[style=numeric,sorting=none, maxnames = 99, minnames = 99]{biblatex}
\AtEveryBibitem{%
  \ifentrytype{article}{%
    \renewbibmacro{in:}{}
  }{}
}

\usepackage{xpatch}             

\title{
DAE-HardNet: A Physics Constrained Neural Network Enforcing Differential-Algebraic Hard Constraints
}

\author{
  Rahul Golder \thanks{Artie McFerrin Department of Chemical Engineering, Texas A\&M University, College Station, TX 77843-3122, USA} \\
  \texttt{rahulgolder8420@tamu.edu}
  \And
  Bimol Nath Roy \footnotemark[1]\\
  \texttt{bimolnathroy@tamu.edu}
  \And
  M. M. Faruque Hasan \footnotemark[1]{ }  \thanks{Texas A\&M Energy Institute, Texas A\&M University, College Station, TX 77843, USA}{ }  \thanks{Corresponding author: \texttt{hasan@tamu.edu}}\\
  \texttt{hasan@tamu.edu} \\
}

\begin{document}
\maketitle

\begin{abstract}
Traditional physics-informed neural networks (PINNs) do not always satisfy physics based constraints, especially when the constraints include differential operators. Rather, they minimize the constraint violations in a soft way. Strict satisfaction of differential-algebraic equations (DAEs) to embed domain knowledge and first-principles in data-driven models is generally challenging. This is because data-driven models consider the original functions to be black-box whose derivatives can only be obtained after evaluating the functions. We introduce \texttt{DAE-HardNet}, a \textit{physics-constrained} (rather than simply \textit{physics-informed}) neural network that learns both the functions and their derivatives simultaneously, while enforcing  algebraic as well as differential constraints. This is done by projecting model predictions onto the constraint manifold using a differentiable projection layer. We apply DAE-HardNet to several systems and test problems governed by DAEs, including the dynamic Lotka-Volterra predator-prey system and transient heat conduction. We also show the ability of DAE-HardNet to estimate unknown parameters through a parameter estimation problem. Compared to  multilayer perceptrons (MLPs) and PINNs, DAE-HardNet achieves orders of magnitude reduction in the physics loss while maintaining the prediction accuracy. It has the added benefits of learning the derivatives which improves the constrained learning of the backbone neural network prior to the projection layer. For specific problems, this suggests that the projection layer can be bypassed for faster inference. The current implementation and codes are available at \hyperlink{https://github.com/SOULS-TAMU/DAE-HardNet}{https://github.com/SOULS-TAMU/DAE-HardNet}.

\end{abstract}


\section{Introduction}
\label{sec:Introduction}
Physical systems are often described with first principles represented by ordinary or partial differential equations (ODEs or PDEs), algebraic equality/inequality constraints. First principles-based models require numerical simulation over a grid which results from a discretized version of the problem. These high fidelity models are computationally intensive and often difficult to generalize for practical use, such as real-time control or large-scale optimization. Physics-informed neural networks (PINNs) \cite{raissi2019physics, karniadakis2021physics} have recently been introduced as a powerful tool for hybrid modeling of physical systems with partial or complete domain knowledge. While traditional feedforward neural networks (NNs) are purely data driven, PINNs enforce physical constraints by adding penalty to the constraint violation in the loss function. There have been several advances \cite{wang2025kolmogorov, li2020fourier, kovachki2021universal, tran2021factorized, lu2019deeponet, abueidda2025deepokan} in the PINN architecture, and PINNs have been used to model transport phenomena \cite{cai2021physics, mao2020physics, cai2021physics_ht}, biological systems \cite{yazdani2020systems, yunduo2025shallow, ahmadi2024ai},  and among others. Although PINNs are generalizable, they suffer from several critical drawbacks:

\begin{enumerate}
    \item With few exceptions \cite{iftakher2025physics, lastrucci2025enforce, amos2017optnet, donti2021dc3}, most PINN architectures only satisfy constraints "softly" via penalizing constraint violations. However, for safety-critical applications, such as batteries \cite{huang2005modeling}, chemical processes \cite{mukherjee2025physics} and nuclear systems \cite{gong2022data, huang2023review}, it is imperative that the physics and operational constraints are "strictly" met. Furthermore, the multiobjective loss function trades prediction accuracy with physical fidelity, but does not guaranty strict satisfaction of the constraints \cite{raissi2019physics, raissi2020hidden, cai2021physics}. Constraint violation can persist, especially when the data is scarce and highly nonlinear \cite{chen2024physics, wang2022and}. This is an issue for applications where there are cascaded surrogate models. Even small violations can accumulate and propagate, undermining the validity of the full system model \cite{ma2022data}.
    \item PINN training is slow to converge \cite{raissi2019physics, raissi2020hidden, cai2021physics, rathore2024challenges}. Mainly due to generally nonconvex physics loss \cite{wang20222}. Benchmark problems like Allen-Cahn \cite{wight2020solving}, Hamilton-Jacobi PDE \cite{liu2022physics} and compressible Navier Stokes \cite{cai2021physics} with nonlinear convective interactions are nonconvex and using these residuals in the loss function makes the unconstrained minimization difficult to converge. 
    \item None of the existing mehtods guarantee "hard" constraint satisfaction for differential-algebraic (DAE) systems. Modeling differential algebraic systems are comparatively harder task than modeling a system of differential equations \cite{petzold1982differential}. Because PINN minimizes the physics loss in a soft manner it tends to diminish the gradient information \cite{wang2021understanding} to reduce the loss. For example, residual of a partial differential equation $\displaystyle\frac{\partial^{2}v}{\partial x^{2}} + \frac{\partial^{2}v}{\partial y^{2}} = 0$ will be lower if both the gradient terms are small which leads to a trivial solution of the system. This is problematic for deep neural networks where the gradients diminish with increasing depth of hidden layers. This so called "spectral bias", in general, is a key issue in PINNs.
    \item Classical PINN performs better for easy parameter regimes but struggles to locate ground truth even with extensive hyperparameter tuning. Techniques such as curriculum regularization and sequence-to-sequence learning tend to reduce the error by orders of magnitude as compared to regular PINN \cite{krishnapriyan2021characterizing}. However, they do not guarantee complete constraint satisfaction.
    \item For constraints with nonconvex terms, the loss function of traditional PINNs becomes nonconvex, as the constraints are added to the loss function. This is problematic for typical gradient discent-based training algorithms.
\end{enumerate}

There is a growing interest in hard-constrained neural networks, which explicitly embed equality and inequality constraints into their architectures. Incorporating such constraints enforces strict adherence to physical laws, helping to regularize the model in data-scarce regimes and reduce overfitting \cite{marquez2017imposing, min2024hard}. Broadly, hard-constrained learning frameworks can be categorized into two main types: predict-and-complete and projection-based methods. Predict-and-complete methods directly predict a subset of outputs and reconstruct the remaining variables analytically or numerically using the constraint equations \cite{beucler2021enforcing, donti2021dc3}. In contrast, projection-based approaches correct unconstrained network outputs by mapping them onto the feasible region defined by the constraints, typically through an Euclidean distance minimization problem \cite{chen2024physics, min2024hard}. Both paradigms effectively maintain constraint satisfaction during training and inference. However, these works have primarily limited to algebraic constraints.\\

Several architectures have integrated differentiable optimization layers within neural networks to impose constraints through the Karush–Kuhn–Tucker (KKT) conditions \cite{nocedal1999numerical}. The OptNet framework \cite{amos2017optnet} first implemented this concept for convex programs using implicit differentiation \cite{agrawal2019differentiable}. These developments allow networks to solve constrained optimization problems within the forward pass while enabling gradient propagation through the solution mapping. Subsequent works such as HardNet-Cvx \cite{min2024hard} and DC3 \cite{donti2021dc3} have extended these ideas to convex and nonlinear problems, employing fixed-point solvers and the implicit function theorem for efficient gradient computation. For systems governed by linear constraints, analytical closed-form projection layers can be derived, removing the need for iterative solvers \cite{chen2021theory, chen2024physics}. Lastrucci and Schweidtmann \cite{lastrucci2025enforce} proposed ENFORCE, a scalable adaptive differentiable projection framework capable of handling sets of $\mathcal{C}^1$ nonlinear equality constraints. Recently, Iftakher et al. developed \texttt{KKT-HardNet} \cite{iftakher2025physics} that enforces both nonlinear equality and inequality constraints using a projection layer based on the KKT system of the distance minimization problem. They handled inequalities using slack variables and complementarity conditions using Fischer-Burmeister reformulation \cite{jiang1997smoothed}. While these frameworks consider strict satisfaction of algebraic constraints, to our best knowledge there remains a need for strict satisfaction of differential constraints.\\

Strict satisfaction of ODE or PDE-based differential constraints is particularly difficult because the gradient information of a data-generated function can be inferred and subsequently hard-restricted only after the function is evaluated. Other challenges include but not limited to spectral bias, uncertain parameter regime, mesh dependencies and reaching trivial solution due to diminishing derivatives in constraining ODE/PDEs in neural network.\\

In this work, we introduce \texttt{DAE-HardNet}, a neural network that enforces differential algebraic constraints in NN models. DAE-Hardnet predicts derivatives that ensure point-wise constraint satisfaction and improves the derivative information through the loss function. To ensure the constraint satisfaction for the predicted derivatives, we use the KKT conditions of a distance minimization problem. Furthermore, to algebraically represent all the output variables with respect to the derivatives, we propose a multiple point neighborhood approximation based on Taylor's expansion. DAE-HardNet satisfies the governing equations or constraints during training and boundary and initial conditions during inference. We demonstrate the proposed framework on illustrative examples of PDEs and systems of ODEs describing real world systems such as reaction kinetics, Lotka Volterra predator-prey system, and a heat conduction problem. Results show that we gain similar predictions and data losses as in MLP models and physics losses better than traditional PINNs, thus outperforming both MLPs and PINNs in terms of overall prediction and reliavility. To summarize, the major contributions of this work are as follows:
\begin{itemize}
    \item We develop \texttt{DAE-HardNet}, a physics constrained neural network framework for hard constraint satisfaction for differential algebraic equation system. We introduce a projection layer based on the KKT system of an algebraic projection problem to predict the derivatives that point-wise satisfy the constraints. 
    \item We convert the original constraints to differential domain with Taylor expansion (1\textsuperscript{st} or 2\textsuperscript{nd} order) and project through the KKT system. Afterwards, we compute the output variable using Taylor approximation based multiple-point approach which ensures plausible prediction of derivatives.
    \item Instead of traditional physics loss used in PINNs, we minimize the Euclidean distance between predicted and automatic differentiation derivatives along with data loss. Unlike the traditional PINNs that suffer from different structured objectives (which can be highly nonconvex sometimes) in the loss function, our proposed loss function is relatively stable due to its convex objective.
    \item We observe that minimizing the derivative loss in fact minimizes the physics loss derived from auto-differentiation. Moreover, as the training progresses, the constraint violation becomes lower than that of the PINNs. 
    \item During training the NN's predictions before and after projection are found to be similar. This leads to the conclusion that during inference we may not need the projection layer, thereby significantly reducing the inference time to close to those of MLP or PINN.
    \item Our framework allows us to consider unknown physical parameters as model parameters and estimating those. Parameter estimation with DAE-HardNet is more powerful because of the synergistic effect of different loss types, pushing each other towards actual or physically meaningful parameter values.
\end{itemize}

\section{Problem Setup}
\label{sec:problem_setup}
In a typical PINN setup, DAEs derived from known first-principles are enforced in a soft manner by including them in the loss function for training. PINN exploits the automatic differentiation (AD) in machine learning \cite{baydin2018automatic} for computing the derivatives and the loss.\\

In this work, we impose the DAEs as hard constraints that preserve the derivative information. We enforce the differential equations by exploiting the Taylor expansion around a predefined neighborhood of any function that completely relies on derivatives. We define the following setup in this context. For any system with unknown field $\boldsymbol{y}(\boldsymbol{x},t)$ with spatial variables $\boldsymbol{x}$ and temporal variable $t$, we have,

\begin{ceqn}
    \begin{equation}\label{eq:general_dae}
    \begin{aligned}
    & \boldsymbol{\mathcal{D}}(\boldsymbol{x},t,\boldsymbol{y},\boldsymbol{\partial}) = \boldsymbol{0},\quad && \boldsymbol{x} \in \Omega, \quad t \in [0,T]\\
    & \boldsymbol{h}(\boldsymbol{x},t,\boldsymbol{y}) = \boldsymbol{0},\quad && \boldsymbol{x} \in \Omega, \quad t \in [0,T]\\
    & \boldsymbol{g}(\boldsymbol{x},t,\boldsymbol{y}) \leq \boldsymbol{0},\quad && \boldsymbol{x} \in \Omega, \quad t \in [0,T]\\
    &\boldsymbol{y}(\boldsymbol{x}, t_0) = \boldsymbol{\mathcal{G}}_{IC}(\boldsymbol{x}), \quad && \boldsymbol{x} \in \Omega\\
    &\boldsymbol{y}(\boldsymbol{x}, t) = \boldsymbol{\mathcal{G}}_{BC}(t), \quad && \boldsymbol{x}\in \partial\Omega, \quad t\in[0,T]
    \end{aligned}
    \end{equation}
\end{ceqn}

where, $\boldsymbol{\mathcal{D}}$ represents the set of differential equations (ordinary or partial), $\mathcal{D}_i(\boldsymbol{x},t,\boldsymbol{y},\boldsymbol{\partial}) = 0$ for $i\in\mathcal{N}_D$ with differential operators $\boldsymbol{\partial}$, $\boldsymbol{h}$ represents the set of algebraic equality constraints $h_j(\boldsymbol{x},t,\boldsymbol{y}) = 0$ for $j\in\mathcal{N}_E$ and $\boldsymbol{g}$ represents the set of inequality constraints $g_k(\boldsymbol{x},t,\boldsymbol{y}) \leq 0$ for $k\in\mathcal{N}_I$, all defining the physics of the system derived from the first-principles. Furthermore, $\boldsymbol{\mathcal{G}}_{IC}$ and $\boldsymbol{\mathcal{G}}_{BC}$ are the initial and boundary conditions, respectively, which can be Dirichlet, Neuman or mixed type. We also have $\Omega \subseteq R^{d}$ as the spatial space, and $\partial\Omega$ as the spatial boundary space. Given a dataset for $\{(\boldsymbol{x}_i,t_i,\bar{\boldsymbol{y}}_i)\}_{i=1}^N$, where, the inputs are $(\boldsymbol{x},t)$ and outputs are $\boldsymbol{\bar{y}}$, we are interested to predict $\bar{\boldsymbol{y}}(\boldsymbol{x},t)$ along with $\bar{\boldsymbol{\partial}}$ while ensuring satisfaction of Eq. \ref{eq:general_dae} for point-wise prediction of $(\boldsymbol{x},t,\bar{\boldsymbol{y}},\bar{\boldsymbol{\partial}})$. All bold symbols represents vector of multiple elements.

\section{DAE-HardNet Framework}
\label{sec:daehardnet-framework}
\begin{figure}[htbp!]
    \centering
    \includegraphics[width=\linewidth]{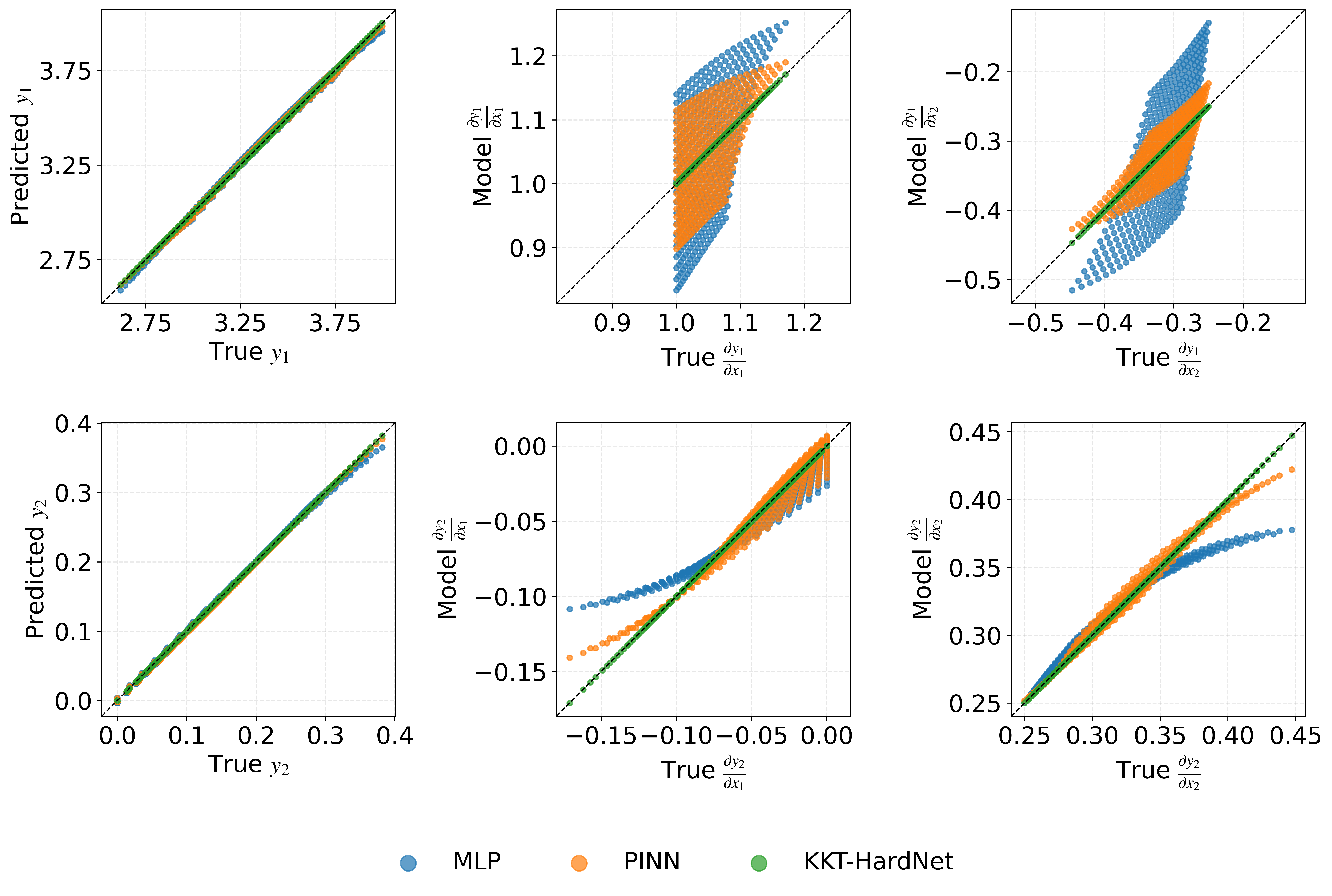}
    \caption{Parity plots for predictions of functions and their derivatives.}
    \label{fig:kkt_hardnet_gradient_parity_plots}
\end{figure}

In most physical systems, the input-output relation is given by completely differential or differential and algebraic equations. In our previous work \cite{iftakher2025physics}, we provided a framework \texttt{KKT-HardNet} for neural network to enforce nonlinear algebraic equality and inequality constraints through KKT-Newton projection. It is worthwhile to illustrate the capability of KKT-HardNet to capture the derivative information compared to MLP and Vanilla-PINN with a small example. We train $f:\mathbb{R}^2\mapsto\mathbb{R}^2$ such that for input $(x_1,x_2)^\top$, the output $(y_1,y_2)^\top$ is the solution of the quadratic system, $Y^2-x_1Y+x_2=0$. Two implicit constraints of this system are $y_1+y_2=x_1$ and $y_1y_2=x_2$. We consider $x_1\in[3,4]$ and $x_2\in[0,1]$ satisfying $x_1^2-4x_2\geq0$ to ensure real output. We provide the parity plot for the output predictions and the first order derivative information in Figure \ref{fig:kkt_hardnet_gradient_parity_plots}. Table \ref{tab:R2_score} provides the $R^2$ values for the prediction and the derivatives. Although the predictions from MLP and PINN are comparable to those of KKT-HardNet (similar $R^2$ values for prediction), the corresponding $R^2$ values for the first order derivative (through auto differentiation) indicates that the derivative information is not preserved always in MLP and PINN, whereas KKT-HardNet preserves the derivative information. This motivates us to extend the projection-based approach using KKT-HardNet architecture to enfore DAEs in DAE-HardNet.

\begin{table}[htbp!]
    \centering
    \caption{Comparison of $R^{2}$ for the predictions of outputs and their derivatives.}
    \label{tab:R2_score}
    \begin{tabular}{lcccccc}
    \toprule
    Model & $\displaystyle y_1$ & $\displaystyle y_2$ &
    $\displaystyle \frac{\partial y_1}{\partial x_1}$ &
    $\displaystyle \frac{\partial y_1}{\partial x_2}$ &
    $\displaystyle \frac{\partial y_2}{\partial x_1}$ &
    $\displaystyle \frac{\partial y_2}{\partial x_2}$ \\ \midrule
    MLP & 0.9981 & 0.9992 & -4.1172 & -1.6285 & 0.8929 & 0.8990 \\
    PINN & 0.9993 & 0.9998 & -1.3728 & 0.6227 & 0.9659 & 0.9864 \\
    KKT-HardNet & 1.0000 & 1.0000 & 1.0000 & 1.0000 & 1.0000 & 1.0000 \\ 
    \bottomrule
    \end{tabular}
\end{table}

\begin{figure}[htbp!]
    \centering
    \includegraphics[width=0.8\linewidth]{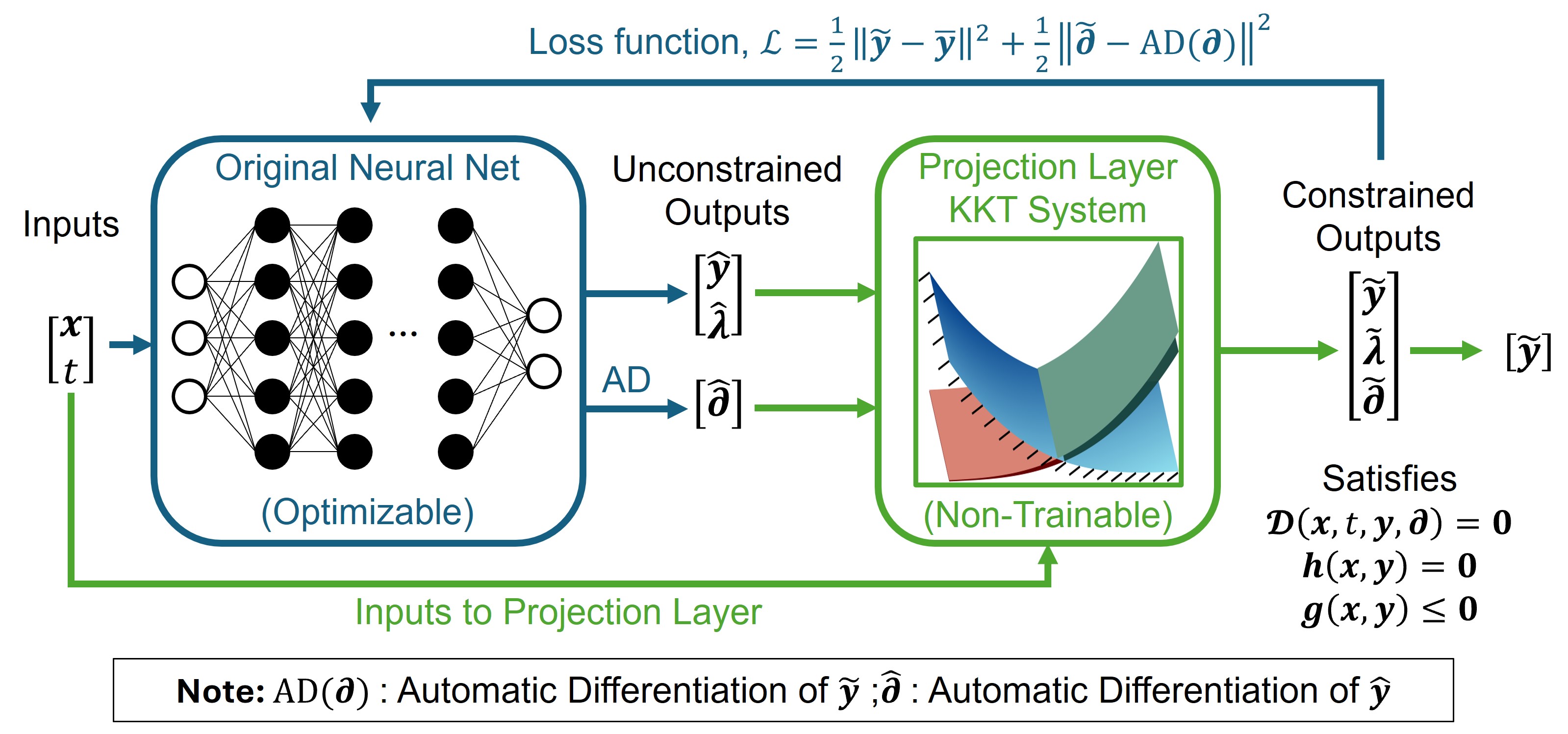}
    \caption{Neural network architecture of DAE-HardNet with exact DAEs satisfaction. Unconstrained outputs including the Lagrangian multipliers are evaluated using a standard neural net (learnable) and the derivative terms are calculated from the network through auto differentiation. The constrained outputs are calculated as the solution of the nonlinear system of equations (non-learnable) corresponding to the KKT relation of a distance minimization problem by considering the derivative terms as independent variables along with Taylor approximation.}
    \label{fig:NNArchitecture}
\end{figure}

To exactly embed domain knowledge that includes differential equations with initial and boundary conditions in learning, in \texttt{DAE-HardNet}, we employ a neural network architecture with two distinguished set of layers that guarantee satisfaction of governing DAEs described in Eq. \ref{eq:general_dae}. Formally, the goal is to train a neural network framework $\boldsymbol{y} = \mathcal{F_{NN}}(\boldsymbol{\Theta}; \boldsymbol{x}, t)$ such that the predicted output $\boldsymbol{y}$ always belongs to the feasible set $\mathcal{S}$, where $\mathcal{S}$ is defined as $\mathcal{S}=\{\boldsymbol{y}|\boldsymbol{\mathcal{D}=0,h=0,g\leq0}\}$. Here, $\boldsymbol{\Theta}$ represents the tunable parameters (weights and biases) of the neural network. Figure \ref{fig:NNArchitecture} shows the visual representation of the proposed architecture. We introduce a KKT-based projection layer after the unconstrained prediction to ensure the feasibility described by the the governing equations. This projection layer introduces the Lagrangian multipliers $\boldsymbol{\lambda}$.\\

Starting with the given spatial and temporal domain we first get the output $\hat{\boldsymbol{Y}}=\mathcal{NN}(\boldsymbol{\Theta}; \boldsymbol{x}, t)$ using a neural network backbone where $\hat{\boldsymbol{Y}}=[\hat{\boldsymbol{y}}\ \ \hat{\boldsymbol{\lambda}}\ \ \hat{\boldsymbol{\partial}}]$, $\hat{\boldsymbol{y}}$ are the soft constrained predictions of the output variables $\boldsymbol{y}$, $\hat{\boldsymbol{\lambda}}$ are the predicted Lagrangian multipliers and $\hat{\boldsymbol{\partial}}$ are the auto differentiation derivatives of $\hat{\boldsymbol{y}}$. We then project $\hat{\boldsymbol{Y}}$ through the projection layer $\mathcal{P}:\hat{\boldsymbol{Y}}\mapsto\tilde{\boldsymbol{Y}}$ where $\tilde{\boldsymbol{Y}}=[\tilde{\boldsymbol{y}}\ \ \tilde{\boldsymbol{\lambda}}\ \ \tilde{\boldsymbol{\partial}}]$ that satisfy Eq. \ref{eq:general_dae}. Finally, we train the model based on the output $\tilde{\boldsymbol{y}}$ and $\tilde{\boldsymbol{\partial}}$.

\subsection{Projection Layer}
\label{subsec:projection_layer}
\begin{figure}
    \centering
    \includegraphics[width=0.7\linewidth]{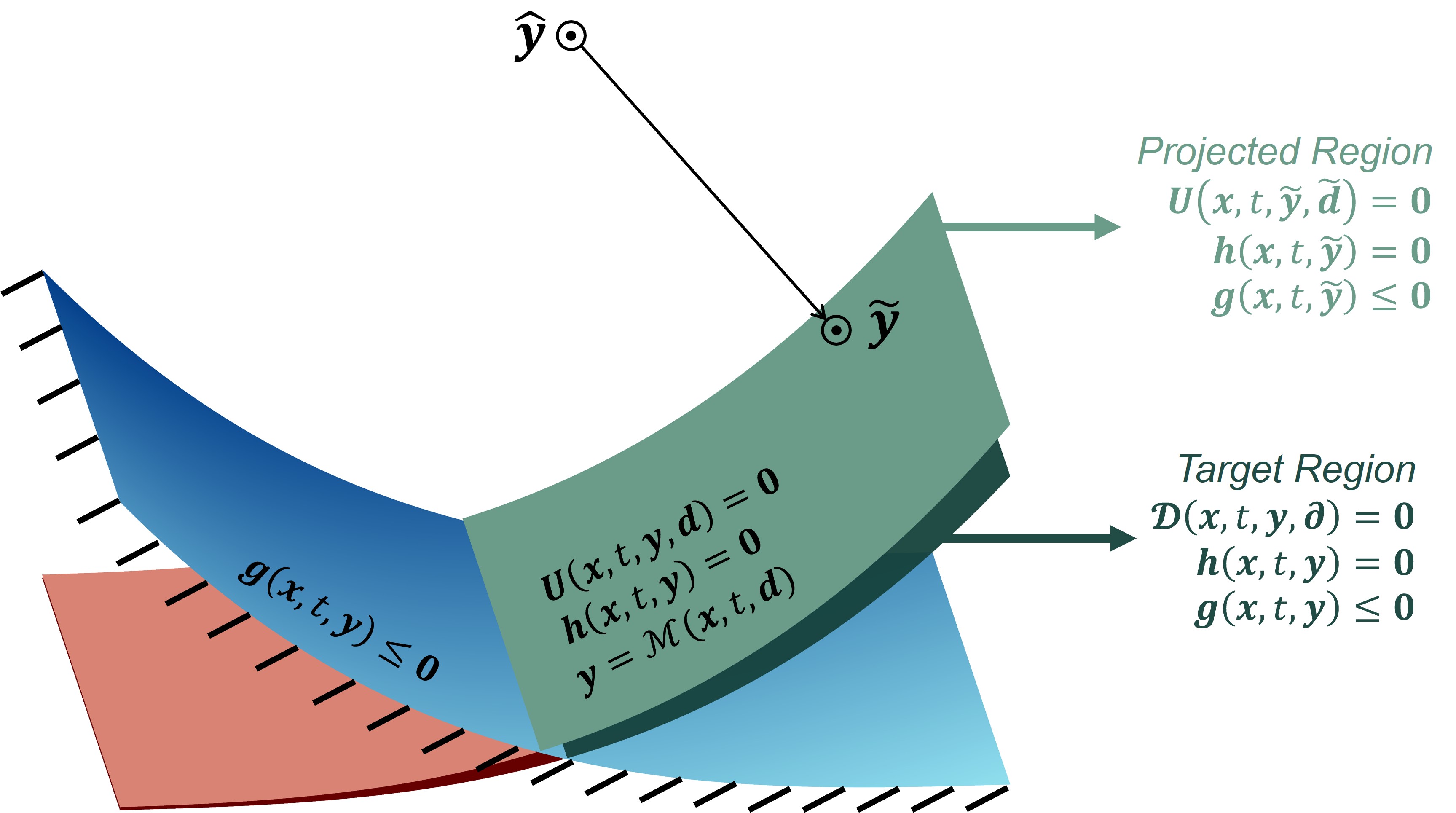}
    \caption{Projection of neural network outputs onto the constraint manifold. The target region is the actual solution of the problem. The difference between target region and projected region reduces with training as the derivative loss approaches zero.}
    \label{fig:projection_layer}
\end{figure}

In our approach, we augment the standard neural network with a projection layer $\mathcal{P}$ that acts as a corrector (an illustration is shown in in Figure \ref{fig:projection_layer}). The projected derivatives satisfy the constraints. The derivative loss term in the loss function (Section \ref{sec:loss_function}) reduces the difference between the target region (actual solution) and the projected region.

The strict enforcement of constraints in the projection layer is ensured by solving the following constrained distance minimization problem:

\begin{ceqn}
    \begin{equation}
    \label{eq-proj-general}
    \begin{aligned}
    \tilde{\boldsymbol{y}} =\; &\arg\ \min_{\boldsymbol{y}} \; \tfrac{1}{2} \left\| \boldsymbol{y} - \hat{\boldsymbol{y}} \right\|^2 \\
    &\quad \text{s.t.} \;\; {\mathcal{D}}_i(\boldsymbol{x},t,\boldsymbol{y},\boldsymbol{\partial}) = {0},\quad \quad &&\forall i\in \mathcal{N}_D,\\
    &\quad\quad\;\; ~~{h}_j(\boldsymbol{x},t,\boldsymbol{y}) = {0},\quad \quad &&\forall j\in \mathcal{N}_E,\\
    &\quad\quad\;\; ~~{g}_k(\boldsymbol{x},t,\boldsymbol{y}) \leq {0},\quad \quad && \forall k\in \mathcal{N}_I.
    \end{aligned}
    \end{equation}
\end{ceqn}

This dynamic optimization problem searches for the minimum distance solution from $\hat{\boldsymbol{y}}$ that lies on the feasible region defined by the DAEs. Solving such problem to optimality is not trivial and computationally expensive. Therefore, we introduce the multivariate Taylor expansion (up to second order) for $\boldsymbol{y}$. For any function $y(\boldsymbol{x},t);\ y:\mathbb{R}^n\times\mathbb{R}_{+}\mapsto \mathbb{R}$, a second order multivariate approximation for $y(\boldsymbol{x},t)$ about the point $(\boldsymbol{x}_0,t_0)$ is defined as follows:

\begin{ceqn}
    \begin{equation}
    \label{eq-taylor-series1}
    \begin{aligned}
    y(\boldsymbol{x},t) \;\approx\; & y(\boldsymbol{x}_0,t_0) + \nabla_{\boldsymbol{x}} y\bigg|_{(\boldsymbol{x}_0,t_0)}^{\top} (\boldsymbol{x}-\boldsymbol{x}_0) + \frac{\partial y}{\partial t}\bigg|_{(\boldsymbol{x}_0,t_0)}\,(t-t_0) + \frac{1}{2} (\boldsymbol{x}-\boldsymbol{x}_0)^{\top} \nabla^2_{\boldsymbol{x}\boldsymbol{x}} y\bigg|_{(\boldsymbol{x}_0,t_0)} (\boldsymbol{x}-\boldsymbol{x}_0) \\
                                    & + (t-t_0)\, \nabla_{\boldsymbol{x}t}^2 y\bigg|_{(\boldsymbol{x}_0,t_0)}^{\top} (\boldsymbol{x}-\boldsymbol{x}_0)+ \frac{1}{2}\frac{\partial^2 y}{\partial t^2}\bigg|_{(\boldsymbol{x}_0,t_0)}\,(t-t_0)^2,
    \end{aligned}
    \end{equation}
\end{ceqn}

where,

\begin{ceqn}
    \begin{equation}
    \label{eq-taylor-series2}
    \begin{gathered}
    \nabla_{\boldsymbol{x}} y\bigg|_{(\boldsymbol{x}_0,t_0)} = \begin{bmatrix}
    \frac{\partial y}{\partial x_1} \\
    \vdots \\
    \frac{\partial y}{\partial x_n}
    \end{bmatrix}_{(\boldsymbol{x}_0,t_0)}
    \in \mathbb{R}^n \quad ; \quad
            \nabla^2_{\boldsymbol{x}t} y\bigg|_{(\boldsymbol{x}_0,t_0)} 
            = \begin{bmatrix}
            \frac{\partial^2 y}{\partial x_1 \partial t}\\
            \vdots \\
            \frac{\partial^2 y}{\partial x_n \partial t}
            \end{bmatrix}_{(\boldsymbol{x}_0,t_0)}
            \in \mathbb{R}^n \quad ; \ \text{and}\\    
    \nabla^2_{\boldsymbol{x}\boldsymbol{x}} y\bigg|_{(\boldsymbol{x}_0,t_0)} 
    = \begin{bmatrix}
    \frac{\partial^2 y}{\partial x_1^2} & \cdots & \frac{\partial^2 y}{\partial x_1 \partial x_n} \\
    \vdots & \ddots & \vdots \\
    \frac{\partial^2 y}{\partial x_n \partial x_1} & \cdots & \frac{\partial^2 y}{\partial x_n^2}
    \end{bmatrix}_{(\boldsymbol{x}_0,t_0)}
    \in \mathbb{R}^{n\times n}.            
    \end{gathered}
    \end{equation}
\end{ceqn}

From here on, we use the notation $\partial_a \;:=\; \frac{\partial y}{\partial a},\;\;\partial_{ab} \;:=\; \frac{\partial^2 y}{\partial a\,\partial b}$, to denote the first- and second-order partial derivatives, respectively. For $n=3$, the Taylor approximation given in Eq. \ref{eq-taylor-series1} can be written as:

\begin{ceqn}
\begin{equation}
\label{eq-taylor-detailed}
\begin{aligned}
y(x_{1,0},x_{2,0},x_{3,0},t_0) \approx\; & y(x_1,x_2,x_3,t) - \partial_{x_1}\bigg|_{0}(x_1-x_{1,0})
- \partial_{x_2}\bigg|_{0}(x_2-x_{2,0})
- \partial_{x_3}\bigg|_{0}(x_3-x_{3,0})\\
& - \partial_{t}\bigg|_{0}(t-t_0) - \frac{1}{2}\partial_{x_1x_1}\bigg|_{0}(x_1-x_{1,0})^2 - \frac{1}{2} \partial_{x_2x_2}\bigg|_{0}(x_2-x_{2,0})^2 \\
& - \frac{1}{2}\partial_{x_3x_3}\bigg|_{0}(x_3-x_{3,0})^2 -\frac{1}{2}\partial_{tt}\bigg|_{0}(t-t_0)^2 - \partial_{x_1x_2}\bigg|_{0}(x_1-x_{1,0})(x_2-x_{2,0}) \\
& - \partial_{x_1x_3}\bigg|_{0}(x_1-x_{1,0})(x_3-x_{3,0}) -\partial_{x_2x_3}\bigg|_{0}(x_2-x_{2,0})(x_3-x_{3,0}) \\
& - \partial_{x_1t}\bigg|_{0}(x_1-x_{1,0})(t-t_0) - \partial_{x_2t}\bigg|_{0}(x_2-x_{2,0})(t-t_0) -\partial_{x_3t}\bigg|_{0}(x_3-x_{3,0})(t-t_0).
\end{aligned}
\end{equation}
\end{ceqn}

To reduce the number of derivative variables which are unlikely to occur in the constraints (cross effect: $\nabla^2_{\boldsymbol{x}t}$ or $\nabla^2_{x_ix_j}$) and to capture the pure curvature in each dimension, we consider \textit{multiple-point neighborhood} approach by evaluating the function at each dimension within a small step, $\Delta$. For $x_1 - x_{1,0} = \Delta$, Eq. \ref{eq-taylor-detailed} becomes,

\begin{ceqn}
\begin{equation}
\label{eq-taylor-x1}
\begin{aligned}
y(x_{1,0},x_{2,0},x_{3,0},t_0) \approx\; & y(x_{1,0}+\Delta,x_{2,0},x_{3,0},t_0) - \partial_{x_1}\bigg|_{0}\Delta - \frac{1}{2}\partial_{x_1x_1}\bigg|_{0}\Delta^2 .
\end{aligned}
\end{equation}
\end{ceqn}

Similarly, for other dimensions, we get

\begin{ceqn}
\begin{equation}
\label{eq-taylor-x2x3t}
\begin{aligned}
y(x_{1,0},x_{2,0},x_{3,0},t_0) \approx\; & y(x_{1,0},x_{2,0}+\Delta,x_{3,0},t_0) - \partial_{x_2}\bigg|_{0}\Delta - \frac{1}{2}\partial_{x_2x_2}\bigg|_{0}\Delta^2\\
y(x_{1,0},x_{2,0},x_{3,0},t_0) \approx\; & y(x_{1,0},x_{2,0},x_{3,0}+\Delta,t_0) - \partial_{x_3}\bigg|_{0}\Delta - \frac{1}{2}\partial_{x_3x_3}\bigg|_{0}\Delta^2\\
y(x_{1,0},x_{2,0},x_{3,0},t_0) \approx\; & y(x_{1,0},x_{2,0},x_{3,0},t_0+\Delta) - \partial_{t}\bigg|_{0}\Delta - \frac{1}{2}\partial_{tt}\bigg|_{0}\Delta^2.
\end{aligned}
\end{equation}
\end{ceqn}

By summing all the approximations from neighborhood,

\begin{ceqn}
\begin{equation}
\label{eq-taylor-x2x3t_new}
\begin{aligned}
y(x_{1,0},x_{2,0},x_{3,0},t_0) \approx\; & \frac{1}{4}\Bigg[y(x_{1,0}+\Delta,x_{2,0},x_{3,0},t_0) + y(x_{1,0},x_{2,0}+\Delta,x_{3,0},t_0) \\
&+ y(x_{1,0},x_{2,0},x_{3,0}+\Delta,t_0) + y(x_{1,0},x_{2,0},x_{3,0},t_0+\Delta)\\ 
&- \Delta\left(\partial_{x_1}\bigg|_{0} + \partial_{x_2}\bigg|_{0} + \partial_{x_3}\bigg|_{0} + \partial_{t}\bigg|_{0}\right)\\
&-\frac{1}{2}\Delta^2\left(\partial_{x_1x_1}\bigg|_{0} + \partial_{x_2x_2}\bigg|_{0} + \partial_{x_3x_3}\bigg|_{0} + \partial_{tt}\bigg|_{0}\right)\Bigg]
\end{aligned}
\end{equation}
\end{ceqn}
The generalized form of \textit{multiple-point neighborhood} (for $X = \{\boldsymbol{x},t\} = \{x_1,x_2,\dots x_n,t\},\lvert X\rvert = n+1$) can be expressed as follows:

\begin{ceqn}
    \begin{equation}
        \label{eq-multiple-point-neighborhood}
        y(\boldsymbol{x},t) \approx \mathcal{M}(\boldsymbol{x},t,\boldsymbol{\partial}) =  \frac{1}{\lvert X\rvert}\Bigg[\sum_{i\in X}y([\boldsymbol{x},t]+\boldsymbol{\Delta}_i) - \Delta \sum_{i\in X}\partial_i - \frac{1}{2}\Delta^2\sum_{i\in X}\partial_{ii}\Bigg]
    \end{equation}
\end{ceqn}

where, $\mathcal{M}$ denotes the multiple-point neighborhood approximation, $X$ is the set of all independent variables and $\boldsymbol{\Delta}_i\in \mathbb{R}^{1\times\lvert X\rvert}$ is the step vector in $i$-th dimension,

\begin{ceqn}
    \begin{equation}
    \boldsymbol{\Delta}_i= \big[\;\delta_{1i}\Delta,\;\delta_{2i}\Delta,\;\ldots,\;\delta_{\lvert X\rvert,i}\Delta\;\big],\quad \text{where,} \quad  
\delta_{ki} \;=\;
\begin{cases}
1, & k=i,\\
0, & k\neq i.
\end{cases}        
    \end{equation}
\end{ceqn}

The optimization problem described in Eq. \ref{eq-proj-general} for $\boldsymbol{x} \in \Omega$ and $t \in [0,T]$ is then reformulated with the approximation given by Eq. \ref{eq-multiple-point-neighborhood} such that

\begin{ceqn}
    \begin{equation}
    \label{eq-proj-taylor}
    \begin{aligned}
    \tilde{\boldsymbol{y}} =\; &\arg\ \min_{\boldsymbol{y}} \; \tfrac{1}{2} \left\| \boldsymbol{y} - \hat{\boldsymbol{y}} \right\|^2 \\
    &\quad \text{s.t.} \;\; {\mathcal{D}}_i(\boldsymbol{x},t,\boldsymbol{y},\boldsymbol{\partial}) = {0} ,\quad \quad && \forall i\in \mathcal{N}_D,\\
    &\quad\quad\;\; ~~{h}_j(\boldsymbol{x},t,\boldsymbol{y}) = {0} ,\quad \quad && \forall j\in \mathcal{N}_E,\\
    &\quad\quad\;\; ~~{g}_k(\boldsymbol{x},t,\boldsymbol{y}) \leq {0} ,\quad \quad && \forall k\in \mathcal{N}_I,\\
    &\quad\quad\;\; ~~{y}_p(\boldsymbol{x},t) = \mathcal{M}_p(\boldsymbol{x},t,\boldsymbol{\partial}), \quad \quad && \forall p\in \mathcal{N}_y,
    \end{aligned}
    \end{equation}
\end{ceqn}

where, $\mathcal{M}_p$ is the multiple-point neighborhood approximation of variable $y_p$, and $\mathcal{N}_y$ represents the set of output variables. This dynamic optimization problem is still infinite dimensional due to the inclusion of differential operators. However, the functional evaluations of $\boldsymbol{y}$ are now related to the point-wise derivatives. We consider the differential operators $\boldsymbol{\partial}$ as individual variables and define them as $\boldsymbol{d}$. We define the algebraic form of the differential equations $\mathcal{D}_i(\boldsymbol{x},t,\boldsymbol{y},\boldsymbol{\partial})$ as $U_i(\boldsymbol{x},t,\boldsymbol{y},\boldsymbol{d})$. Now the optimization problem in Eq. \ref{eq-proj-taylor} becomes a nonlinear program (NLP) with algebraic equality and inequality constraints.

\begin{ceqn}
    \begin{equation}
    \label{eq-proj-nlp}
    \begin{aligned}
    \tilde{\boldsymbol{y}}, \tilde{\boldsymbol{d}} =\; &\arg\min_{\boldsymbol{y}, \boldsymbol{d}} \; \tfrac{1}{2} \left\| \boldsymbol{y} - \hat{\boldsymbol{y}} \right\|^2 \\
    &\quad \text{s.t.} \;\; {U}_i(\boldsymbol{x},t,\boldsymbol{y},\boldsymbol{d}) = {0} ,\quad \quad && \forall i\in \mathcal{N}_D,\\
    &\quad\quad\;\; ~~{h}_j(\boldsymbol{x},t,\boldsymbol{y}) = {0} ,\quad \quad && \forall j\in \mathcal{N}_E,\\
    &\quad\quad\;\; ~~{g}_k(\boldsymbol{x},t,\boldsymbol{y}) \leq {0} ,\quad \quad && \forall k\in \mathcal{N}_I,\\
    &\quad\quad\;\; ~~{y}_p(\boldsymbol{x},t) = \mathcal{M}_p(\boldsymbol{x},t,\boldsymbol{d}), \quad \quad && \forall p\in \mathcal{N}_y.
    \end{aligned}
    \end{equation}
\end{ceqn}

By defining the Lagrangian multipliers $\lambda_i^D\in\mathbb{R}\ , \lambda_j^E\in\mathbb{R}\ , \lambda_k^I \geq0 \ ,\lambda_p\in\mathbb{R}$ and non-negative slack $s_k$ for $k\in\mathcal{N}_I$, we can write the KKT conditions of the problem in Eq. \ref{eq-proj-nlp} as follows:

\begin{ceqn}
\begin{equation}
    \begin{aligned}
        &\text{(Stationary)} \quad \quad &&\boldsymbol{y}-\hat{\boldsymbol{y}} +\sum_{i \in \mathcal{N}_D}\lambda_i^{D}\,\nabla_{(\boldsymbol{y},\boldsymbol{d})}U_i(\boldsymbol{x},t,\boldsymbol{y},\boldsymbol{d}) + \sum_{j \in \mathcal{N}_E}\lambda_k^{E}\,\nabla_{(\boldsymbol{y},\boldsymbol{d})}{h}_j(\boldsymbol{x},t,\boldsymbol{y}) \\
        &                       &&+ \sum_{k \in \mathcal{N}_I}\lambda_k^{I}\,\nabla_{(\boldsymbol{y},\boldsymbol{d})}g_k(\boldsymbol{x},\boldsymbol{y}) + \sum_{p \in \mathcal{N}_y} \lambda_p \, \nabla_{(\boldsymbol{y} , \boldsymbol{d})}\left[{y}_p(\boldsymbol{x},t) - \mathcal{M}_p(\boldsymbol{x},t,\boldsymbol{d})\right]= {0} \\[4pt]
        &\text{(Primal )}\quad \quad && {U}_i(\boldsymbol{x},t,\boldsymbol{y},\boldsymbol{d}) = {0} ,\quad \quad \forall i\in \mathcal{N}_D\\[4pt]
        &\text{(Primal )}\quad \quad && {h}_j(\boldsymbol{x},t,\boldsymbol{y}) = {0} ,\quad \quad \forall j\in \mathcal{N}_E\\[4pt]
        &\text{(Primal )}\quad \quad  && {g}_k(\boldsymbol{x},t,\boldsymbol{y}) + s_k = {0} ,\quad \quad \forall k\in \mathcal{N}_I \\[4pt]
        &\text{(Primal )}\quad \quad  && {y}_p(\boldsymbol{x},t) - \mathcal{M}_p(\boldsymbol{x},t,\boldsymbol{d}) = 0, \quad \quad \forall p\in \mathcal{N}_y \\[4pt]
        &\text{(Dual + Complementary)}\quad &&\lambda_k^I + s_k - \sqrt{(\lambda_k^{I})^2 + s_k^2}\;=\; 0, \quad \quad k \in \mathcal{N}_I. \\[4pt]
  \label{eq:kktconditions}
    \end{aligned}
\end{equation}
\end{ceqn}

Note that the dual feasibility and complementarity conditions are reformulated via the exact Fischer-Burmeister reformulation \cite{jiang1997smoothed}. The first-order KKT conditions given in Eq. \ref{eq:kktconditions} is a square system of nonlinear equations. We refer the readers to our previous work, KKT-HardNet \cite{iftakher2025physics}, for the detailed formulation strategy to find the square system of equations for any general nonlinear optimization problem from the KKT conditions through the Fischer-Burmeister reformulation. We define this system of nonlinear equation as the projection layer $\mathcal{P}$ which can be referred as an implicit function $\mathcal{P}(\boldsymbol{x},t,\boldsymbol{\hat{y}},\boldsymbol{y},\boldsymbol{d},\boldsymbol{\lambda})$ where, $\boldsymbol{x}\ , t$ are the inputs, $\boldsymbol{\hat{y}}$ is the neural network backbone prediction, $\boldsymbol{y}$ is the projected output, $\boldsymbol{d}$ is the projected derivative, and $\boldsymbol{\lambda} = [\boldsymbol{\lambda}^D \ \boldsymbol{\lambda}^E \ \boldsymbol{\lambda}^I \ \boldsymbol{\lambda}^p \ \boldsymbol{s}]$ are the projected Lagrangian and slack variables. We are particularly interested in the values of the projected $\boldsymbol{y}$ and projected $\boldsymbol{d}$. We define $\tilde{\boldsymbol{y}},\ \tilde{\boldsymbol{d}}$ as the solution from the projection layer.\\

\textit{\textbf{Remark 1}} One could proceed without Taylor approximation and declare the derivative terms ($\boldsymbol{\partial}$) as independent variables together with $\boldsymbol{y}$ in formulating the projection layer. However, since $\boldsymbol{\partial}$ are not inherently linked to $\boldsymbol{y}$, this introduces degrees of freedom in the model. This motivates the need for coupling between $\boldsymbol{y}$ and $\boldsymbol{\partial}$ which is achieved with Taylor approximation based \textit{multiple-point neighborhood} approach. If the differential equations are ordinary differential equations, then the \textit{multiple-point neighborhood} approximation becomes same as the original Taylor approximation.\\

\textit{\textbf{Remark 2}} Note that, in our implementation, we evaluate $y([\boldsymbol{x},t]+\boldsymbol{\Delta}_i) \ \forall i \in X $ from the neural network backbone before the projection layer. The approximation is exact when $\Delta$ approaches zero $(\Delta \to 0)$. However, a very small value of $\Delta$ is expected to make the system unstable. Therefore, we take a reasonable value to avoid instability in the model.

\subsection{Loss Function}
\label{sec:loss_function}
While the KKT-HardNet framework uses a loss function to minimize loss between data and constrained prediction as follows:

\begin{ceqn}
\begin{equation}
\label{eq-los-KKTHN}
\mathcal{L}_{\texttt{KKT-Hardnet}} = MSE\left(\tilde{\boldsymbol{y}}, \bar{\boldsymbol{y}}\right)
\end{equation}
\end{ceqn}

we propose the following loss function in the DAE-HardNet framework with a tuning parameter $\omega$:

\begin{ceqn}
\begin{equation}
\label{eq-los-DAEHN}
\mathcal{L}_{\texttt{DAE-Hardnet}} = MSE(\tilde{\boldsymbol{y}}, \bar{\boldsymbol{y}}) + \omega \times MSE(\tilde{\boldsymbol{d}}, AD(\boldsymbol{\partial})) ,
\end{equation}
\end{ceqn}

where, $MSE$ refers to mean squared error and $AD$ is the corresponding derivative value ($\partial$) of $\tilde{\boldsymbol{y}}$ evaluated using automatic differentiation (AutoGrad) feature in PyTorch library \cite{paszke2019pytorch}. The additional second term with a weighting parameter multiplied with it guides the derivatives to match the predicted derivatives which satisfy the constraints. Solving the KKT system only guarantees a stationary point. However, the data loss (first term in Eq. \ref{eq-los-DAEHN}) in the loss function acts as the driving force for shifting the stationary point projection to the data, and the derivative loss (second term in Eq. \ref{eq-los-DAEHN}) improves the overall model's derivative information. The improved derivative information has a synergistic effect on the predictions ($\tilde{\boldsymbol{y}}$). Moreover, both the loss terms are similar (target based) in the proposed loss function which makes it more attractive from an optimization stand point. 

It is worthwhile to mention, due to the structure of the loss function, the training is not unstable like vanilla PINNs for different weights. Moreover, the loss landscape becomes increasingly complex in regular PINN \cite{krishnapriyan2021characterizing} due to different structured loss multiplied by a regularization factor. This makes the problem harder to optimize. In our proposed loss function, the structure is an ellipsoidal norm ball which can be optimized in a stable way. The ellipsoidal norm ball becomes a spherical norm ball when the regularization $\omega$ is 1. This resolves the issue of handling a complex loss landscape.

\subsection{Training}
\begin{figure}[htbp!]
    \centering
    \includegraphics[width=\linewidth]{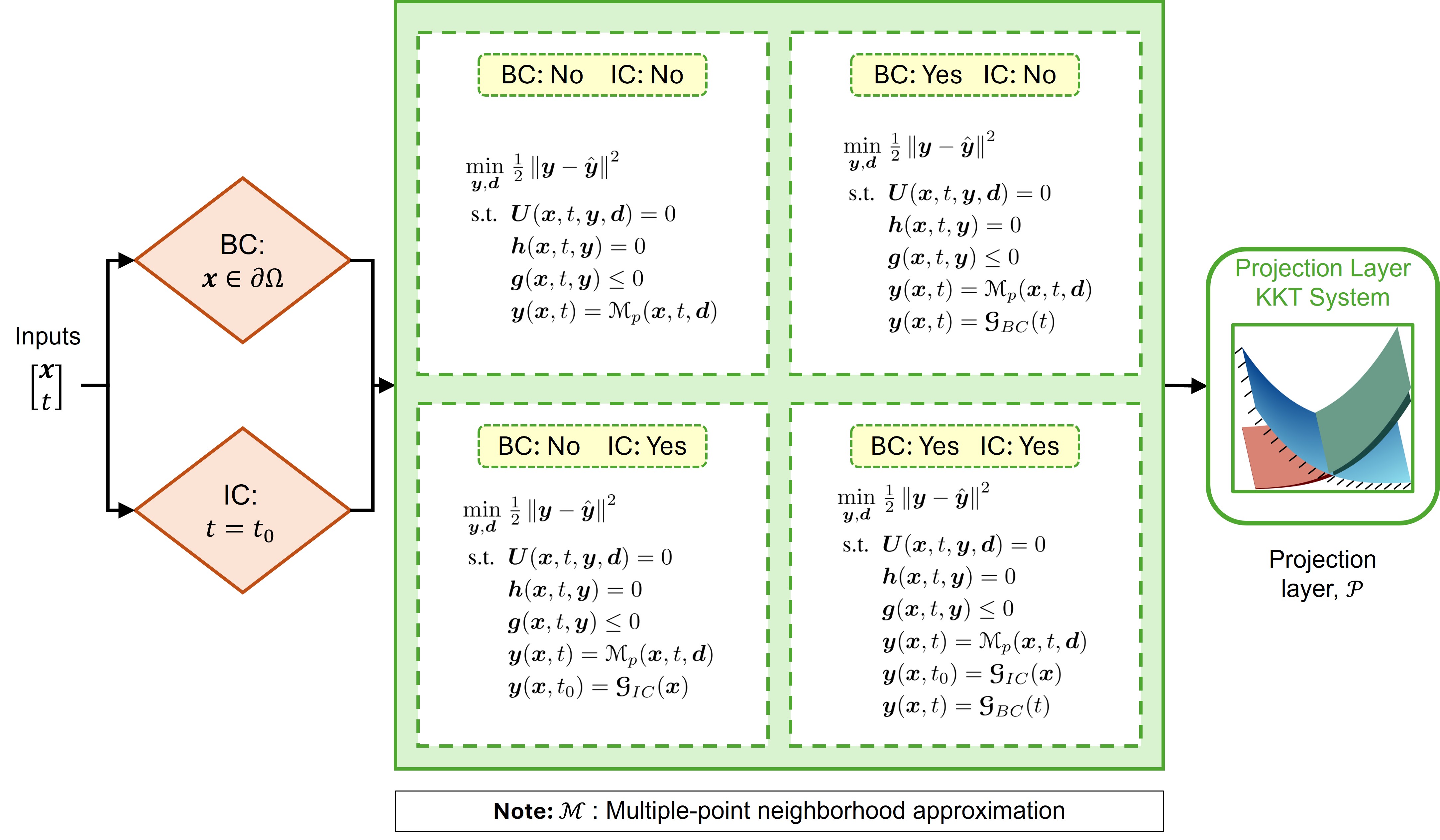}
    \caption{The pool of KKT systems created before model training due to different combination of the input features. During training \textit{BC: No, IC: No} condition is imposed ensuring satisfaction of the governing equations during training for all points. Meanwhile during inference based on the input feature, corresponding KKT system is selected as the projection layer which satisfies the corresponding BC/IC if applicable.}
    \label{fig:KKT_combination}
\end{figure}

Until this point, we addressed learning the derivative space and training the model accordingly. In our proposed framework we do not include the boundary and initial conditions in the projection during training. However, during inference based on the inputs, we include the boundary and initial conditions that apply as constraints in Eq. \ref{eq-proj-taylor}. The corresponding KKT projection layer is then used to compute the outputs which also satisfy the initial and boundary conditions. A visual representation of this is provided in Figure \ref{fig:KKT_combination}. The overall training and inference of DAE-HardNet are described in Algorithm \ref{alg:DAE-Hardnet}.

\begin{algorithm}[htbp!]
\caption{DAE-Hardnet: Hard Differential Algebraic Constraint Satisfaction via Differentiable Projection}
\label{alg:DAE-Hardnet}
\begin{algorithmic}[1]
\Require Dataset $\boldsymbol{\mathcal D}=\{(\boldsymbol x, t, \boldsymbol{\bar{y}} \}$; neural net backbone $\mathcal{NN}(\boldsymbol{\Theta};\boldsymbol{x},t):\mathbb R^{m}\!\to\!\mathbb R^{p}$; projection routine $\mathcal{P}(\cdot)$; Constraints as $\boldsymbol{\mathcal{D}}(\boldsymbol{x},t,\boldsymbol{y},\boldsymbol{\partial}) = \boldsymbol{0}$, $\boldsymbol{h}(\boldsymbol{x},t,\boldsymbol{y}) = \boldsymbol{0}$, $\boldsymbol{g}(\boldsymbol{x},t,\boldsymbol{y}) \leq \boldsymbol{0}$; Boundary conditions as $\boldsymbol{y}(\boldsymbol{x}, t) = \boldsymbol{\mathcal{G}}_{BC}(t) \boldsymbol{x}\in \partial\Omega, \quad t\in[0,T]$; Initial conditions as $\boldsymbol{y}(\boldsymbol{x}, t_0) = \boldsymbol{\mathcal{G}}_{IC}(\boldsymbol{x}), \boldsymbol{x} \in \Omega$; the KKT system corresponding to the projection problem (Fig. \ref{fig:KKT_combination}).

\vspace{0.6em}

\Procedure{Train}{$\boldsymbol{\mathcal D}$}
  \State initialize neural network $\mathcal{NN}(\boldsymbol{\Theta};\boldsymbol{x},t)$
  \While{not converged}
    \For{batch $\boldsymbol{\mathcal B}\subset\boldsymbol{\mathcal D}$}
      \For{$(\boldsymbol{x},t, \boldsymbol{\bar{y}})\in \boldsymbol{\mathcal B}$}
        \State \textbf{predict (unconstrained):} $\hat{\boldsymbol{y}}, \hat{\boldsymbol{\lambda}}\gets \mathcal{NN}(\boldsymbol{\Theta};\boldsymbol{x},t)$
        \State \textbf{compute:} $\hat{\boldsymbol{\partial}}$ using automatic differentiation
        \State \textbf{project:} $\tilde{\boldsymbol{y}},\tilde{\boldsymbol{\lambda}},\tilde{\boldsymbol{d}}\gets\mathcal{P}(\hat{\boldsymbol{y}}, \hat{\boldsymbol{\lambda}}, \hat{\boldsymbol{\partial}})$
        \State \textbf{compute derivatives:} $AD(\boldsymbol{\partial})$ using automatic differentiation
        \State \textbf{compute loss:} $\mathcal{L}=MSE(\tilde{\boldsymbol{y}}, \bar{\boldsymbol{y}}) + MSE(\tilde{\boldsymbol{d}}, AD(\boldsymbol{\partial}))$ 
      \EndFor
      \State \textbf{update} $\boldsymbol{\Theta}$ using $\boldsymbol{\nabla_\Theta} \mathcal{L}$ by backpropagation
    \EndFor
  \EndWhile
\EndProcedure
\vspace{0.4em}

\Procedure{Test}{$\boldsymbol x, t, \mathcal{NN}(\boldsymbol{\Theta};\boldsymbol{x},t)$}
  \State $\hat{\boldsymbol{y}}, \hat{\boldsymbol{\lambda}}\gets \mathcal{NN}(\boldsymbol{\Theta};\boldsymbol{x},t)$
  \State \textbf{select projection layer:} $\mathcal{P}(\cdot)$ based on the input $\boldsymbol{x},t$
  \State $\tilde{\boldsymbol{y}},\tilde{\boldsymbol{\lambda}},\tilde{\boldsymbol{d}}\gets\mathcal{P}(\hat{\boldsymbol{y}}, \hat{\boldsymbol{\lambda}}, \hat{\boldsymbol{\partial}})$
  \State \Return $\tilde{\boldsymbol y}$
\EndProcedure
\vspace{0.6em}
\end{algorithmic}
\end{algorithm}

\section{Numerical experiments}
\label{sec:numerical_experiments}
We present six examples involving various DAE systems of practical importance to illustrate the capability of DAE-HardNet to enforce these DAE constraints on the training and inference. In all examples, the data were split to use $80\%$ and $20\%$ for training and validation respectively. We trained all the models with PyTorch package using Adam optimizer \cite{kingma2014adam}. All numerical experiments were performed on a 13\textsuperscript{th}~Gen Intel\textsuperscript{\textregistered} Core\textsuperscript{\texttrademark} i7--13700 (24-core CPU) with 32 GB RAM running Linux OS. The model implementation are made available on our GitHub repository at: \hyperlink{https://github.com/SOULS-TAMU/DAE-HardNet}{https://github.com/SOULS-TAMU/DAE-HardNet}.\\

We provide all the tunable and user defined model parameters for all experiments including \texttt{num\_epochs}: total number of epochs used for training, \texttt{mdoel\_depth}: number of hidden layers in the NN backbone, \texttt{hidden\_dim}: number of nodes at each hidden layer, \texttt{lr}: learning rate for the Adam optimizer, \texttt{num\_points}: number of data points used for training the model, \texttt{pinn\_reg\_factor}: regularization factor used in PINN model, \texttt{hardnet\_reg\_factor}: regularization factor used in DAE-HardNet model, \texttt{taylor\_offset}: value of $\Delta$ used in the Taylor approximation, \texttt{taylor\_order}: order of Taylor approximation, \texttt{eta}: achieved value of the testing data loss after which the projection layer activates, \texttt{newton\_step\_length}: step length for Newton projection layer, \texttt{max\_newton\_iter}: maximum number of iterations given for newton projection layer, \texttt{noise\_std}: standard deviation of the noise if added, \texttt{noise\_mean}: mean of the applied noise and \texttt{noise\_scale}: scale of the added noise. To assess the predictive performance and physical consistency of the models, we report the following metrics: mean squared error (MSE), root mean squared error (RMSE), and mean absolute constraint violation. We provide the MSE derivative loss between predicted and model derivatives for DAE-HardNet model.

For predictions $\tilde{\boldsymbol y}_i\in\mathbb{R}^p$ and targets $\bar{\boldsymbol y}_i\in\mathbb{R}^p$, we use

\begin{align*}
\mathrm{MSE} &= \frac{1}{N p}\sum_{i=1}^N \sum_{j=1}^p \big(\tilde{y}_{ij}-\bar{y}_{ij}\big)^2 \\
\text{and}, \quad \mathrm{RMSE} &= \sqrt{\mathrm{MSE}}.
\end{align*}

Mean absolute constraint violation is given by
\begin{equation*}
\label{eq:mean_violation}
\mathrm{Violation}
\;=\;
\frac{1}{N\,m}\sum_{i=1}^N
\left(
\sum_{k\in\mathcal N_D} \big|\,\mathcal{D}_k(\boldsymbol x_i,t_i,\tilde{\boldsymbol y}_i,\tilde{\boldsymbol{\partial}}_i)\,\big|
\;+\;
\sum_{k\in\mathcal N_E} \big|\,h_k(\boldsymbol x_i,t_i, \tilde{\boldsymbol y}_i)\,\big|
\;+\;
\sum_{\ell\in\mathcal N_I} \text{ReLU}\big(\,g_\ell(\boldsymbol x_i, t_i, \tilde{\boldsymbol y}_i)\,\big)
\right)
\end{equation*}

where, $m = |\mathcal N_E| + |\mathcal N_I| + |\mathcal N_D|$, and $\text{ReLU}(a)=\max\{a,0\}$.

\textit{\textbf{Remark 3}} We include the initial and boundary conditions after training, which restricts the framework to apply from solving systems of ODEs or PDEs. With the current approach, Dirichlet, Neumann or mixed type of boundary conditions can be included in the KKT system. Other boundary conditions such as periodic boundary conditions are not supported in the current implementation.

\subsection{Example 1: System of ODEs}
Our first example is a system of ODEs \cite{greenberg1998advanced} given by,
\begin{equation}
\label{example:system-of-odes}
\begin{aligned}
    \frac{d^{2}y_{1}}{dx_1^{2}} - y_{1} + 2y_{2} &= 0, \\[6pt]
    \frac{d^{2}y_{2}}{dx_1^{2}} + 4y_{2} - 2y_{1} + x_1^{2} - 1 &= 0.
\end{aligned}
\end{equation}

The general solution of the system is as follows,

\begin{equation}
\label{example:system-of-odes-sol}
\begin{aligned}
    y_{1}(t) &= \frac{1}{18}x_1^{4} - \frac{5}{9}x_1^{2} - \frac{8}{27} 
                + A\sin(\sqrt{3}x_1) + B\cos(\sqrt{3}x_1) + 2C + 2Dx_1, \\[6pt]
    y_{2}(x_1) &= -\frac{11}{18}x_1^{4} - \frac{1}{36}x_1^{2} - \frac{11}{27} 
                + 2A\sin(\sqrt{3}x_1) + 2B\cos(\sqrt{3}x_1) + C + Dx_1.
\end{aligned}
\end{equation}

where $A,B,C,D$ are constants which are determined by the specific boundary conditions imposed. We have considered the following values for the constants to generate the data: $A = -2.0, \ B = -2.0, \ C = 0.5,\ D = 0.0.$



\begin{figure}[htbp!]
    \centering
    \includegraphics[width=0.9\linewidth]{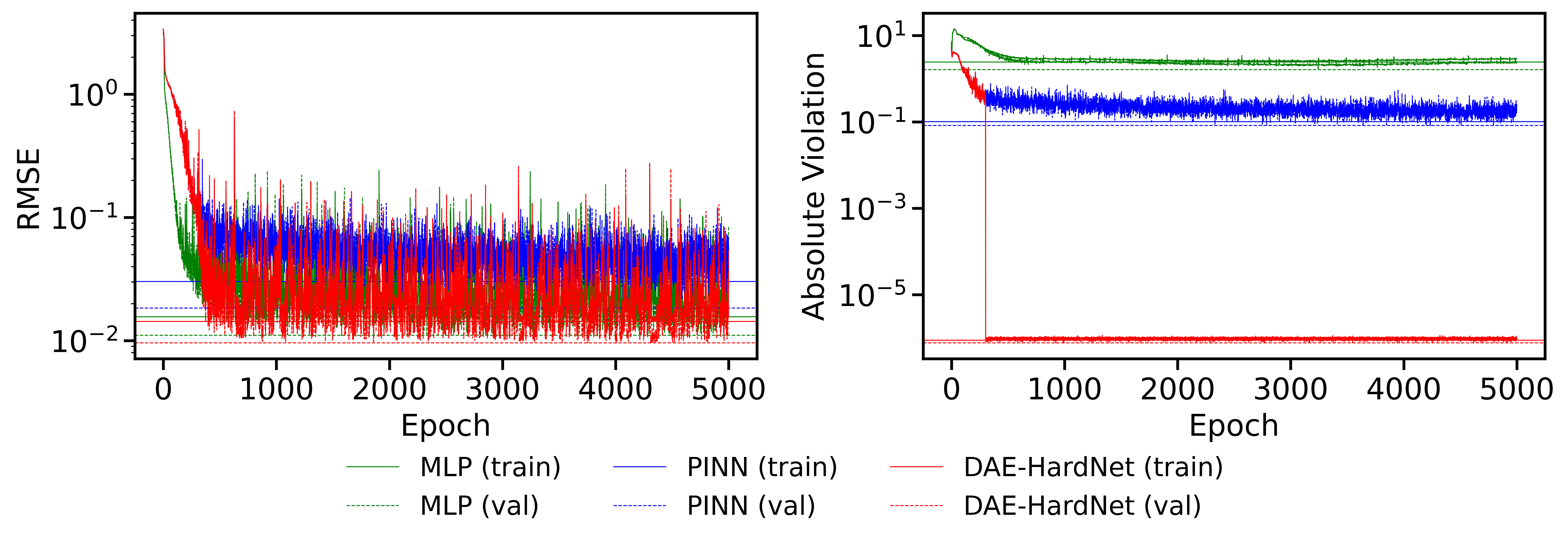}
    \caption{Learning curves for the system of ODEs in Example 1. Left: MSE data loss; Right: absolute constraint violation on the projected gradients over 5000 epochs.}
    \label{fig:system_of_ode_training}
\end{figure}

\begin{figure}[htbp!]
    \centering
    \includegraphics[width=0.8\linewidth]{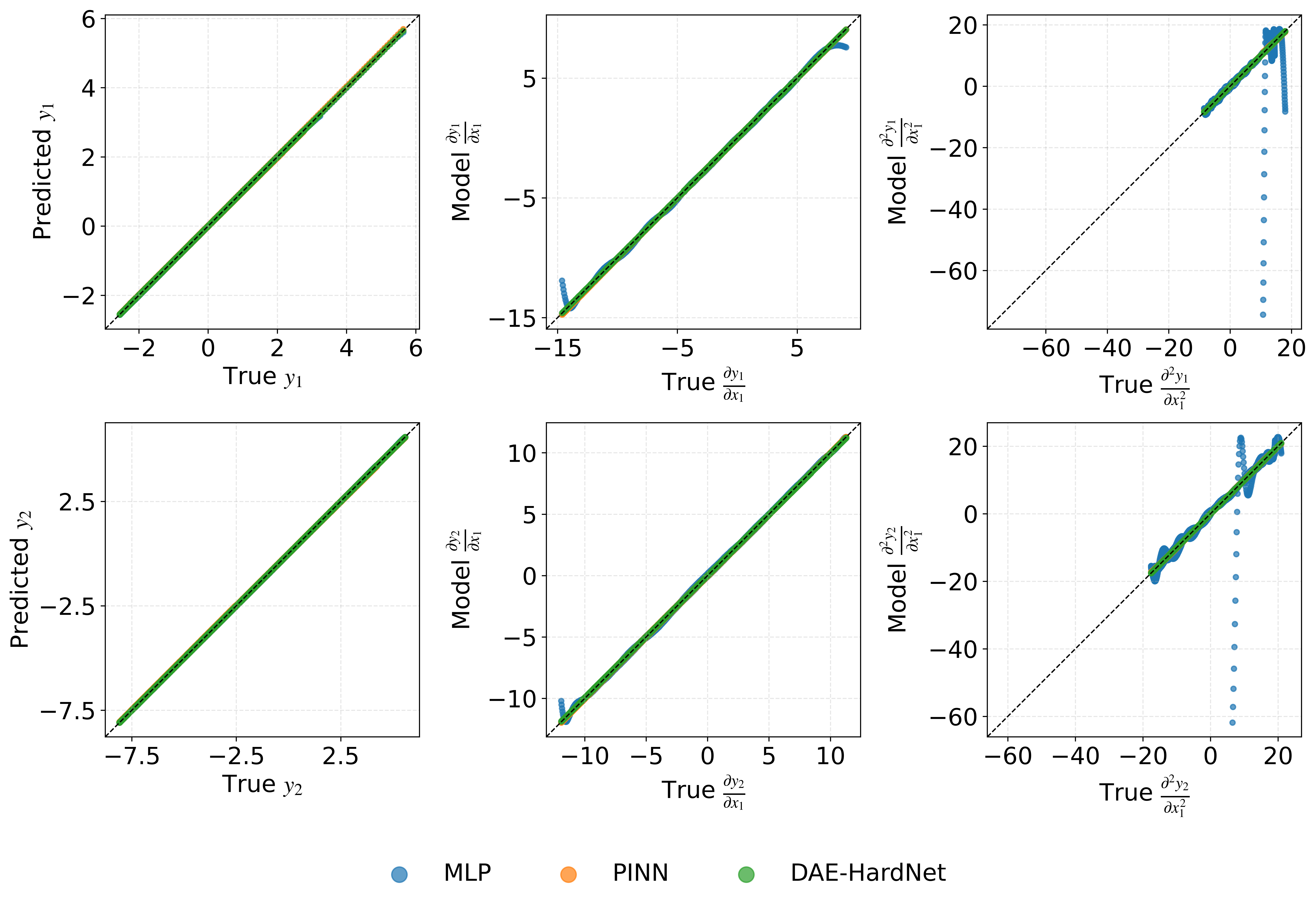}
    \caption{Model Derivative for MLP and PINN and predicted derivative for DAE-HardNet vs. the actual derivative values for the system of odes in Example 1. Based on the predicted derivative the loss was minimized and the model was optimized.}
    \label{fig:system_of_ode_gradient_parity_plots}
\end{figure}

The data were generated for a spatial grid of $x_1 \in [-4,4]$. As mentioned in the training algorithm, we activated the projection layer dynamically once the data loss of the PINN training was $0.01$. This gives a heuristically better guess for the projection layer for finding a stationary point. We enforce thetwo ODEs and compared the performance of DAE-HardNet with PINN and MLP, all using identical architecture and optimization settings. The learning curves for MSE and absolute constraint violation are provided in Figure \ref{fig:system_of_ode_training}. Table \ref{tab:systemofodes} provides the MSE data loss, MSE derivative loss and absolute violation for both training and testing for all models along with the model configurations. Derivative loss of order $10^{-4}$ was achieved for DAE-HardNet. Lower MSE data loss was also achieved for DAE-HardNet than that of MLP and PINN for both training and testing. Figure \ref{fig:system_of_ode_gradient_parity_plots} provides visual comparison of the predicted/model derivatives compared to that of the true derivative values. The parity plots shows the accurate prediction of derivatives based on which the DAE-Hardnet model is optimized. 

\begin{table}[htbp!]
    \centering
    \caption{Regression accuracy, constraint enforcement comparison and model configuration for the system of ODEs in illustrative example 1.}
    \label{tab:systemofodes}
    \begin{adjustbox}{width=\textwidth}
    \begin{tabular}{lccccccc}
    \toprule
    \multicolumn{1}{c}{\multirow{2}{*}{\textbf{Model}}} & \multicolumn{3}{c}{\textbf{Training}}                             & \multicolumn{3}{c}{\textbf{Validation}}                           & \multicolumn{1}{c}{\multirow{2}{*}{\textbf{Best Epoch}}} \\ 
    \cmidrule(lr){2-4}\cmidrule(lr){5-7}
    \multicolumn{1}{c}{}    & MSE (Data)                    & MSE (Derivative)          & Abs Violation                 & MSE (Data)                    & MSE (Derivative)          & Abs Violation                 & \multicolumn{1}{c}{}\\ \midrule
    MLP                     & \(3.46\!\times\!10^{-4}\)     & --                        & \(2.60\!\times\!10^{0}\)      & \(1.27\!\times\!10^{-4}\)     & --                        & \(2.10\!\times\!10^{0}\)      & 2850 \\
    PINN                    & \(1.74\!\times\!10^{-3}\)     & --                        & \(1.60\!\times\!10^{-1}\)     & \(4.47\!\times\!10^{-4}\)     & --                        & \(1.00\!\times\!10^{-1}\)     & 4120 \\
    DAEHN                   & \(2.26\!\times\!10^{-4}\)     & \(3.74\!\times\!10^{-4}\) & \(9.45\!\times\!10^{-7}\)     & \(9.71\!\times\!10^{-5}\)     & \(3.11\!\times\!10^{-4}\) & \(9.04\!\times\!10^{-7}\)     & 3150 \\ \midrule
    \multicolumn{8}{c}{\textbf{Model Configurations}}                                                                                                                                                                      \\ \midrule
    \multicolumn{2}{l}{\texttt{num\_epochs}}                & \multicolumn{2}{l}{5000}                                  & \multicolumn{2}{l}{\texttt{taylor\_order}}                & \multicolumn{2}{l}{1}                \\
    \multicolumn{2}{l}{\texttt{model\_depth}}               & \multicolumn{2}{l}{4}                                     & \multicolumn{2}{l}{\texttt{eta}}                          & \multicolumn{2}{l}{0.01}             \\
    \multicolumn{2}{l}{\texttt{hidden\_dim}}                & \multicolumn{2}{l}{32}                                    & \multicolumn{2}{l}{\texttt{newton\_step\_length}}         & \multicolumn{2}{l}{1.00}             \\
    \multicolumn{2}{l}{\texttt{lr}}                         & \multicolumn{2}{l}{0.001}                                 & \multicolumn{2}{l}{\texttt{max\_newton\_iter}}            & \multicolumn{2}{l}{10}               \\
    \multicolumn{2}{l}{\texttt{num\_points}}                & \multicolumn{2}{l}{1500 out of 1500}                      & \multicolumn{2}{l}{\texttt{noise\_std}}                   & \multicolumn{2}{l}{1.00}             \\
    \multicolumn{2}{l}{\texttt{pinn\_reg\_factor}}          & \multicolumn{2}{l}{1.00}                                  & \multicolumn{2}{l}{\texttt{noise\_mean}}                  & \multicolumn{2}{l}{0.00}             \\
    \multicolumn{2}{l}{\texttt{hardnet\_reg\_factor}}       & \multicolumn{2}{l}{1.00}                                  & \multicolumn{2}{l}{\texttt{noise\_scale}}                 & \multicolumn{2}{l}{0.01}             \\ 
    \multicolumn{2}{l}{\texttt{taylor\_offset}}             & \multicolumn{2}{l}{0.10}                                  & \multicolumn{2}{l}{\texttt{}}                             & \multicolumn{2}{l}{}                 \\ \bottomrule
    \end{tabular}
    \end{adjustbox}
\end{table}

\subsection{Example 2: Differential Algebraic Systems describing Oxidation of Carbon monoxide}
This example illustrates the use of DAE-HardNet for systems where both differential and algebraic constraints are present. We consider a catalytic reaction process for the oxidation of $\mathrm{CO}$ by $\mathrm{O}_{2}$ over Bismuth Uranate ($\mathrm{Bi}_{2}\mathrm{UO}_{6}$) \cite{collette1987potential}. The overall reaction is given by,

\begin{equation}
\label{case:DAE}
    \mathrm{CO} + \frac{1}{2}\mathrm{O}_{2} \to \mathrm{CO}_{2}.
\end{equation}

The proposed mechanism of the reaction is given by,

\begin{equation}
\label{case:DAE_Mechanism}
\begin{aligned}
    \mathrm{CO} + \sigma 
    &\xrightleftharpoons[k_{-1}]{k_1} 
    \mathrm{CO}\!-\!\sigma \\[6pt]
    \mathrm{CO}\!-\!\sigma + [\mathrm{O}] 
    &\xrightarrow{k_2} 
    \mathrm{CO_2} + [\cdot]_O + \sigma \\[6pt]
    2[\cdot]_O + \mathrm{O_2} 
    &\xrightarrow{k_3} 
    2[\mathrm{O}]
\end{aligned}
\end{equation}

where $[\mathrm{O}]$ represents the lattice oxygen atom and $[\cdot]_O$ represents an oxygen vacancy at the surface. From the experimental study \cite{collette1987potential}, the reaction rate data was obtained at $400^{\circ}\mathrm{C}$ over $0.5$ g of catalyst. The lattice oxygen atoms were in large excess and equilibrium with respect to adsorption step was established very quickly. At low partial pressure of $\mathrm{CO}$, the system follows a first order kinetics but at high partial pressure of $\mathrm{CO}$ follows a zero order kinetics. Considering this observation the following mathematical model is proposed to define the kinetics of the system:

\begin{equation}
\label{case:DAE_equation}
    -\frac{dP_{\mathrm{CO}}}{dt} = k'\,\theta_{\mathrm{CO}}; \quad \quad
    \theta_{\mathrm{CO}} = k_{1} P_{\mathrm{CO}} \theta_{V}; \quad \quad
    \theta_{V} + \theta_{\mathrm{CO}} = 1
\end{equation}

where $k' = k_{2}P_{\mathrm{O}}$ and since $P_{\mathrm{O}}$ is in large excess $k'$ is considered a constant. 

\begin{figure}[htbp!]
    \centering
    \includegraphics[width=0.9\linewidth]{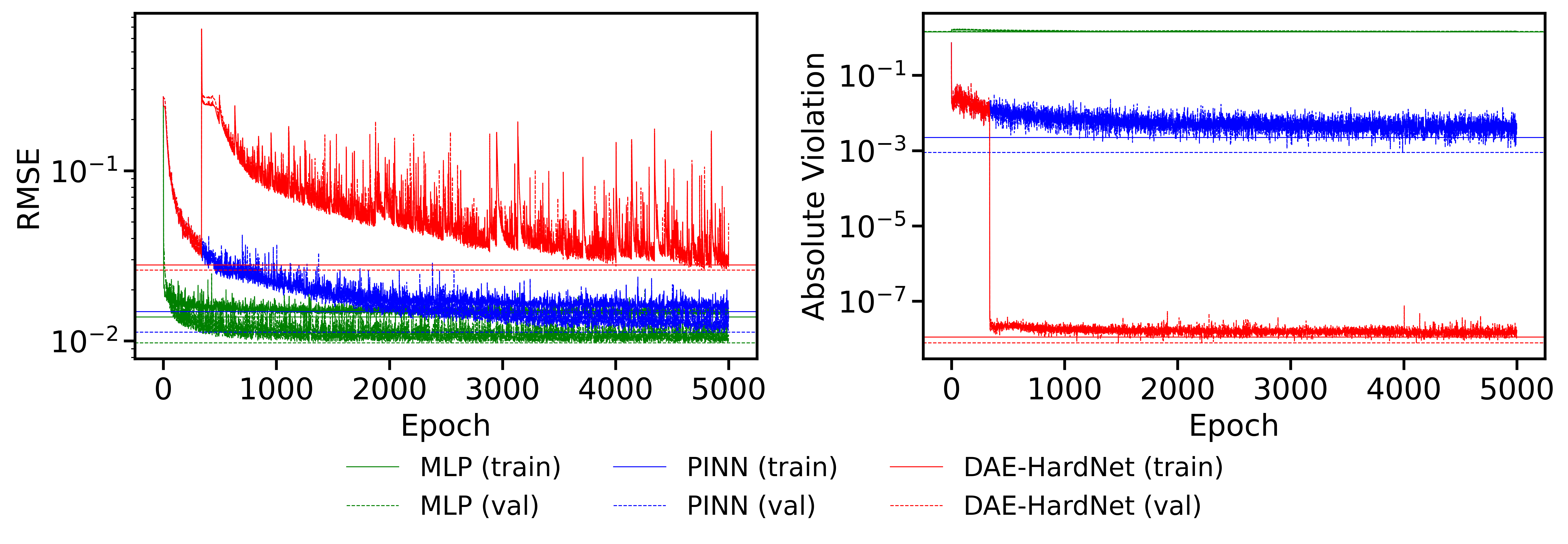}
    \caption{Learning curves for the DAE system in Example 2. Left: MSE data loss; Right: absolute constraint violation on the projected gradients over 5000 epochs.}
    \label{fig:dae_catalysis_training}
\end{figure}


We simulated the given DAE system given by Eq. \ref{case:DAE_equation}. We use a temporal range of $t\in[0,1000]$ to generate the data for training and validation. We activated the projection layer dynamically once the data loss of PINN testing was $0.001$, providing a better guess for projection layer at the beginning. We enforced one ODEs and two algebraic equation and compared the performance of our framework with MLP and PINN. The model architecture and optimization settings were identical for all three models. The learning curves and absolute violation for both training and testing are provided in Figure \ref{fig:dae_catalysis_training}. Table \ref{tab:DAEresults} provides the training metrics and the model configurations. Derivative losses of \(8.03\!\times\!10^{-4}\) and \(1.57\!\times\!10^{-2}\) are achieved during training and testing, respectively. We observed absolute violation of constraints for projected derivatives close to $10^{-9}$.

\begin{table}[htbp!]
    \centering
    \caption{Regression accuracy, constraint enforcement comparison and model configuration for the example of oxidation of $\textrm{CO}$ by $\textrm{O}_2$.}
    \label{tab:DAEresults}
    \begin{adjustbox}{width=\textwidth}
    \begin{tabular}{lccccccc}
    \toprule
    \multicolumn{1}{c}{\multirow{2}{*}{\textbf{Model}}} & \multicolumn{3}{c}{\textbf{Training}}                             & \multicolumn{3}{c}{\textbf{Validation}}                           & \multicolumn{1}{c}{\multirow{2}{*}{\textbf{Best Epoch}}} \\ 
    \cmidrule(lr){2-4}\cmidrule(lr){5-7}
    \multicolumn{1}{c}{}    & MSE (Data)                    & MSE (Derivative)          & Abs Violation                 & MSE (Data)                    & MSE (Derivative)          & Abs Violation                 & \multicolumn{1}{c}{}\\ \midrule
    MLP                     & \(2.25\!\times\!10^{-4}\)     & --                        & \(1.42\!\times\!10^{0}\)      & \(9.56\!\times\!10^{-5}\)     & --                        & \(1.47\!\times\!10^{0}\)      & 3560 \\
    PINN                    & \(2.49\!\times\!10^{-4}\)     & --                        & \(4.68\!\times\!10^{-3}\)     & \(1.29\!\times\!10^{-4}\)     & --                        & \(3.85\!\times\!10^{-3}\)     & 4920 \\
    DAEHN                   & \(8.03\!\times\!10^{-4}\)     & \(9.16\!\times\!10^{-4}\) & \(1.33\!\times\!10^{-8}\)     & \(6.79\!\times\!10^{-4}\)     & \(1.57\!\times\!10^{-2}\) & \(9.34\!\times\!10^{-9}\)     & 4760 \\ \midrule
    \multicolumn{8}{c}{\textbf{Model Configurations}}                                                                                                                                                                      \\ \midrule
    \multicolumn{2}{l}{\texttt{num\_epochs}}                & \multicolumn{2}{l}{5000}                                  & \multicolumn{2}{l}{\texttt{taylor\_order}}                & \multicolumn{2}{l}{1}                \\
    \multicolumn{2}{l}{\texttt{model\_depth}}               & \multicolumn{2}{l}{4}                                     & \multicolumn{2}{l}{\texttt{eta}}                          & \multicolumn{2}{l}{0.001}             \\
    \multicolumn{2}{l}{\texttt{hidden\_dim}}                & \multicolumn{2}{l}{32}                                    & \multicolumn{2}{l}{\texttt{newton\_step\_length}}         & \multicolumn{2}{l}{1.00}             \\
    \multicolumn{2}{l}{\texttt{lr}}                         & \multicolumn{2}{l}{0.001}                                 & \multicolumn{2}{l}{\texttt{max\_newton\_iter}}            & \multicolumn{2}{l}{10}               \\
    \multicolumn{2}{l}{\texttt{num\_points}}                & \multicolumn{2}{l}{2000 out of 2000}                      & \multicolumn{2}{l}{\texttt{noise\_std}}                   & \multicolumn{2}{l}{1.00}             \\
    \multicolumn{2}{l}{\texttt{pinn\_reg\_factor}}          & \multicolumn{2}{l}{1.00}                                  & \multicolumn{2}{l}{\texttt{noise\_mean}}                  & \multicolumn{2}{l}{0.00}             \\
    \multicolumn{2}{l}{\texttt{hardnet\_reg\_factor}}       & \multicolumn{2}{l}{1.00}                                  & \multicolumn{2}{l}{\texttt{noise\_scale}}                 & \multicolumn{2}{l}{0.01}             \\
    \multicolumn{2}{l}{\texttt{taylor\_offset}}              & \multicolumn{2}{l}{0.01}                                  & \multicolumn{2}{l}{\texttt{}}                             & \multicolumn{2}{l}{}                 \\ \bottomrule
    \end{tabular}
    \end{adjustbox}
\end{table}
\subsection{Example 3: Lotka Volterra Predator-Prey System}
\label{case_study:lotka_volterra}
Lotka Volterra Predator Prey model \cite{wangersky1978lotka} is a benchmark example of a complex interaction system. The system is defined by,

\begin{equation}
\label{eq:lotka_volterra_system}
\begin{aligned}
    \frac{dx}{dt} = \alpha x - \beta xy, \\
    \frac{dy}{dt} = -\gamma y + \delta xy
\end{aligned}
\end{equation}

where $x$ and $y$ are the population densities of prey and predator, respectively. Furthermore, $\frac{dx}{dt}$, $\frac{dy}{dt}$ represent the instantaneous growth rates of two populations. The parameters $\alpha$ and $\beta$ describe the maximum per capita growth rate of the prey and the effect of the presence of predators on the prey death rate, respectively. The parameters $\gamma$ and $\delta$ describe the predators per capita death rate, and the effect of the presence of prey on the predator's growth rate, respectively.

\begin{figure}[htbp!]
    \centering
    \includegraphics[width=0.9\linewidth]{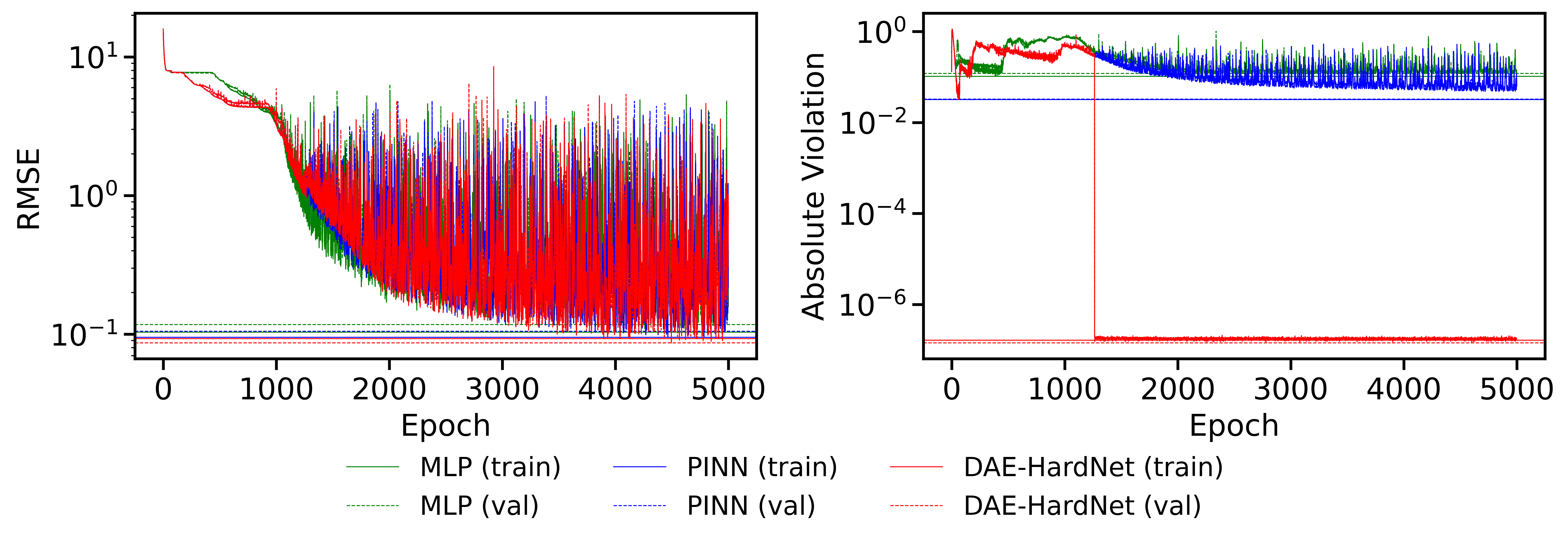}
    \caption{Learning curves for the predator-prey system in Example 3. Left: MSE data loss; Right: absolute constraint violation on the projected gradients over 5000 epochs.}
    \label{fig:lotka_volterra_training}
\end{figure}

We considered initial prey and predator population as $x_{0} = 10$ and $y_{0} = 10$, respectively, with the parameter values as $\alpha=0.1$, $\beta=0.02$, $\gamma=0.4$, $\delta=0.02$. We simulated the system to generate a total of $2000$ temporal data points for both the predator and the prey populations. Figure \ref{fig:lotka_volterra_training} shows the RMSE data loss and absolute violation for projected gradient during training and testing. Table \ref{tab:LotkaVolterra} provides the evaluation metrics for the best fitted epoch and the model configurations for all three models. The derivative loss was found to be $3.65\times10^{-4}$ and $6.83\times10^{-4}$ for training and testing respectively. Moreover the data loss for DAE-HardNet was found to be lower than those of MLP and PINN for both training and testing.


\begin{table}[htbp!]
    \centering
    \caption{Regression accuracy, constraint enforcement comparison and model configuration for the Lotka Volterra predator-prey system.}
    \label{tab:LotkaVolterra}
    \begin{adjustbox}{width=\textwidth}
    \begin{tabular}{lccccccc}
    \toprule
    \multicolumn{1}{c}{\multirow{2}{*}{\textbf{Model}}} & \multicolumn{3}{c}{\textbf{Training}}                             & \multicolumn{3}{c}{\textbf{Validation}}                           & \multicolumn{1}{c}{\multirow{2}{*}{\textbf{Best Epoch}}} \\ 
    \cmidrule(lr){2-4}\cmidrule(lr){5-7}
    \multicolumn{1}{c}{}    & MSE (Data)                    & MSE (Derivative)          & Abs Violation                 & MSE (Data)                    & MSE (Derivative)          & Abs Violation                 & \multicolumn{1}{c}{}\\ \midrule
    MLP                     & \(2.17\!\times\!10^{-2}\)     & --                        & \(1.20\!\times\!10^{-1}\)     & \(1.67\!\times\!10^{-2}\)     & --                        & \(1.30\!\times\!10^{-1}\)     & 4260 \\
    PINN                    & \(1.97\!\times\!10^{-2}\)     & --                        & \(4.97\!\times\!10^{-2}\)     & \(1.10\!\times\!10^{-2}\)     & --                        & \(5.80\!\times\!10^{-2}\)     & 4780 \\
    DAEHN                   & \(1.06\!\times\!10^{-2}\)     & \(3.65\!\times\!10^{-4}\) & \(1.81\!\times\!10^{-7}\)     & \(9.24\!\times\!10^{-3}\)     & \(6.83\!\times\!10^{-4}\) & \(1.83\!\times\!10^{-7}\)     & 4550 \\ \midrule
    \multicolumn{8}{c}{\textbf{Model Configurations}}                                                                                                                                                                      \\ \midrule
    \multicolumn{2}{l}{\texttt{num\_epochs}}                & \multicolumn{2}{l}{5000}                                  & \multicolumn{2}{l}{\texttt{taylor\_order}}                & \multicolumn{2}{l}{1}                \\
    \multicolumn{2}{l}{\texttt{model\_depth}}               & \multicolumn{2}{l}{4}                                     & \multicolumn{2}{l}{\texttt{eta}}                          & \multicolumn{2}{l}{1.00}             \\
    \multicolumn{2}{l}{\texttt{hidden\_dim}}                & \multicolumn{2}{l}{32}                                    & \multicolumn{2}{l}{\texttt{newton\_step\_length}}         & \multicolumn{2}{l}{1.00}             \\
    \multicolumn{2}{l}{\texttt{lr}}                         & \multicolumn{2}{l}{0.001}                                 & \multicolumn{2}{l}{\texttt{max\_newton\_iter}}            & \multicolumn{2}{l}{10}               \\
    \multicolumn{2}{l}{\texttt{num\_points}}                & \multicolumn{2}{l}{2000 out of 2000}                      & \multicolumn{2}{l}{\texttt{noise\_std}}                   & \multicolumn{2}{l}{--}             \\
    \multicolumn{2}{l}{\texttt{pinn\_reg\_factor}}          & \multicolumn{2}{l}{1.00}                                  & \multicolumn{2}{l}{\texttt{noise\_mean}}                  & \multicolumn{2}{l}{--}             \\
    \multicolumn{2}{l}{\texttt{hardnet\_reg\_factor}}       & \multicolumn{2}{l}{100.0}                                 & \multicolumn{2}{l}{\texttt{noise\_scale}}                 & \multicolumn{2}{l}{--}             \\
    \multicolumn{2}{l}{\texttt{taylor\_offset}}              & \multicolumn{2}{l}{0.1}                                  & \multicolumn{2}{l}{\texttt{}}                             & \multicolumn{2}{l}{}                 \\ \bottomrule
    \end{tabular}
    \end{adjustbox}
\end{table}

\subsection{Example 4: Parameter Estimation}
\begin{figure}[htbp!]
    \centering
    \includegraphics[width=\linewidth]{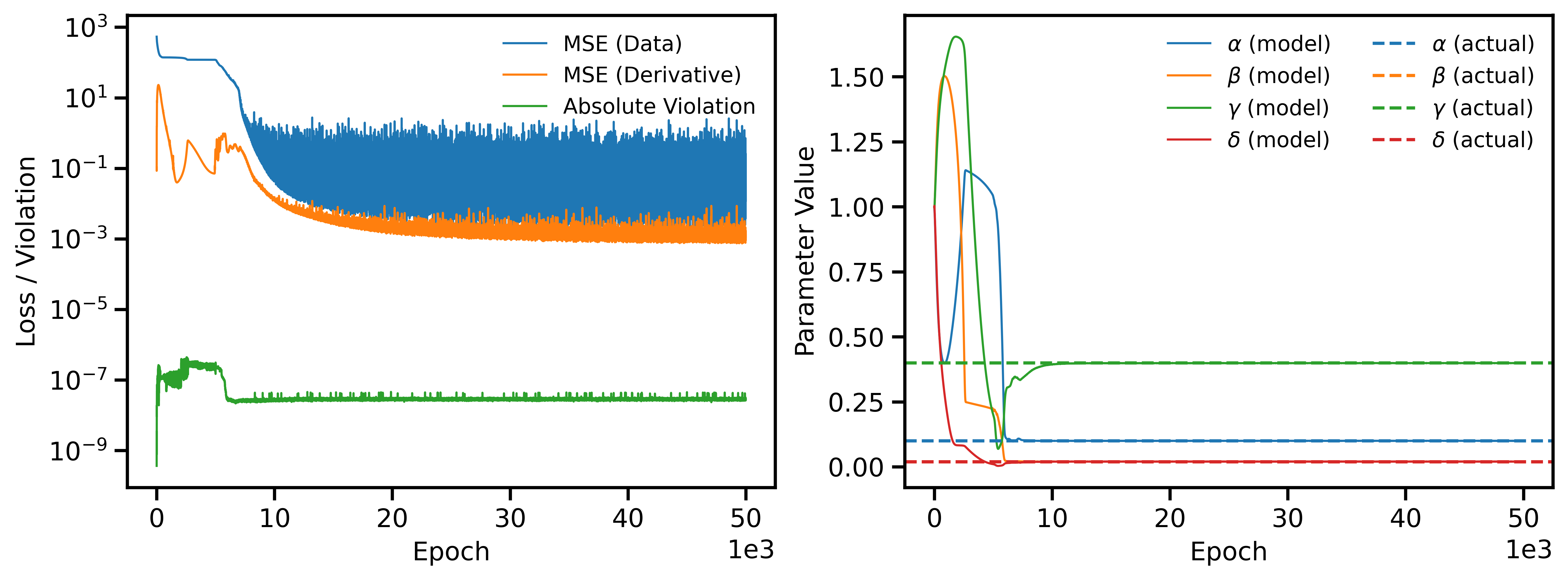}
    \caption{Results for Parameter Estimation. Left: Loss and Violation Propagation during model training. Both Data loss and Derivative loss are shown in the plot. Additionally the violation of the model on the governing equation is also shown. Right: The propagation of the unknown model parameter values are shown for $\alpha, \beta, \gamma$ and $\delta$ respectively.}
    \label{fig:parameter_estimation}
\end{figure}

DAE-HardNet can also be used to perform inverse modeling to estimate unknown parameters of a known system. For this analysis we consider the Lotka-Volterra system for which we have already shown the performance of the model for forward modeling in section \ref{case_study:lotka_volterra}. The system for this problem remains the same as before, but we consider that the unknown properties are $\alpha, \beta, \gamma, \delta$. We consider these parameters as model parameters like weights and biases that the model can update during backpropagation to minimize both data loss and derivative loss. We consider the same configuration of the model provided in section \ref{case_study:lotka_volterra} and train the model for $50000$ epochs. Figure \ref{fig:parameter_estimation} shows the results of parameter estimation. Left plot shows the losses during model training along with governing equation violation of the model. We observe a synergistic effect between data loss and derivative loss because derivative loss trains the model to learn the gradient information effectively which also improves the model's prediction. Additionally, the violation of the model are always in the order of $10^{-7}$ since the projection layer always ensures that the projected values of both the variables and gradient terms satisfies the governing equation. The plot on the right shows the values of model parameters through the epochs. The actual values of $\alpha, \beta, \gamma, \delta$  are given by $0.1, 0.02, 0.4, 0.02$, which is used to generate the data. The figure shows that the model converges to the original parameter values in $10^4$ number of epochs and there is no instability in the training following convergence. This shows that the DAE-HardNet model can be used for solving inverse problems to find unknown parameters of a system.

\subsection{Example 5: PDE with multiple solutions}

\begin{figure}[htbp!]
    \centering
    \includegraphics[width=\linewidth]{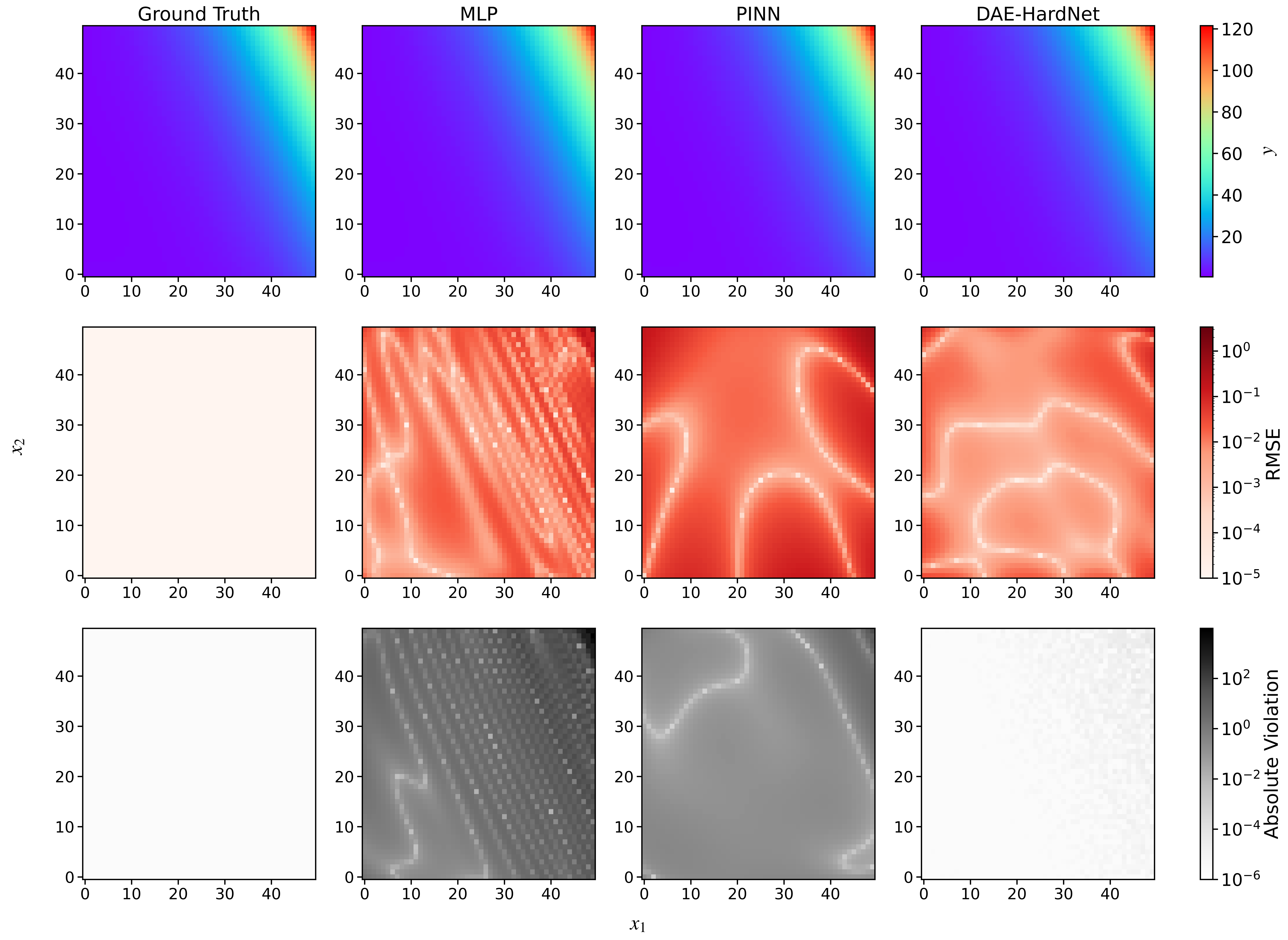}
    \caption{Model prediction comparison for all models with ground truth. Top row: Model prediction heatmap; Middle row: Heatmap for the RMSE loss showing better performance of DAE-HardNet than MLP and PINN; Bottom row: Absolute violation heatmap with the model derivatives for MLP and PINN and with projected derivatives for DAE-HardNet.}
    \label{fig:greenberg_pde_prediction}
\end{figure}

We consider the following second order PDE \cite{greenberg1998advanced} with nonlinearity in both input and output

\begin{ceqn}
\begin{equation}
\label{case:toy_pde}
    \frac{\partial^{2} y_1}{\partial x_1^{2}} - 5 \frac{\partial y_1}{\partial x_2} + y_1 = x_1x_2^{2}(x_2 - 15).
\end{equation}    
\end{ceqn}

The example PDE is of second order and admits the solutions 

\begin{ceqn}
\begin{subequations}
\label{eq:analytical_soln_greenberg}
\begin{align}
y_1(x_1,x_2) = 6\text{e}^{2x_1+x_2}+x_1x_2^3, \label{eq:greeberg_sol1}\\
y_1(x_1,x_2) = -\text{e}^{3x_1+2x_2}+x_1x_2^3 \label{eq:greenberg_sol2} 
\end{align}
\end{subequations}    
\end{ceqn}

\begin{figure}[htbp!]
    \centering
    \includegraphics[width=0.9\linewidth]{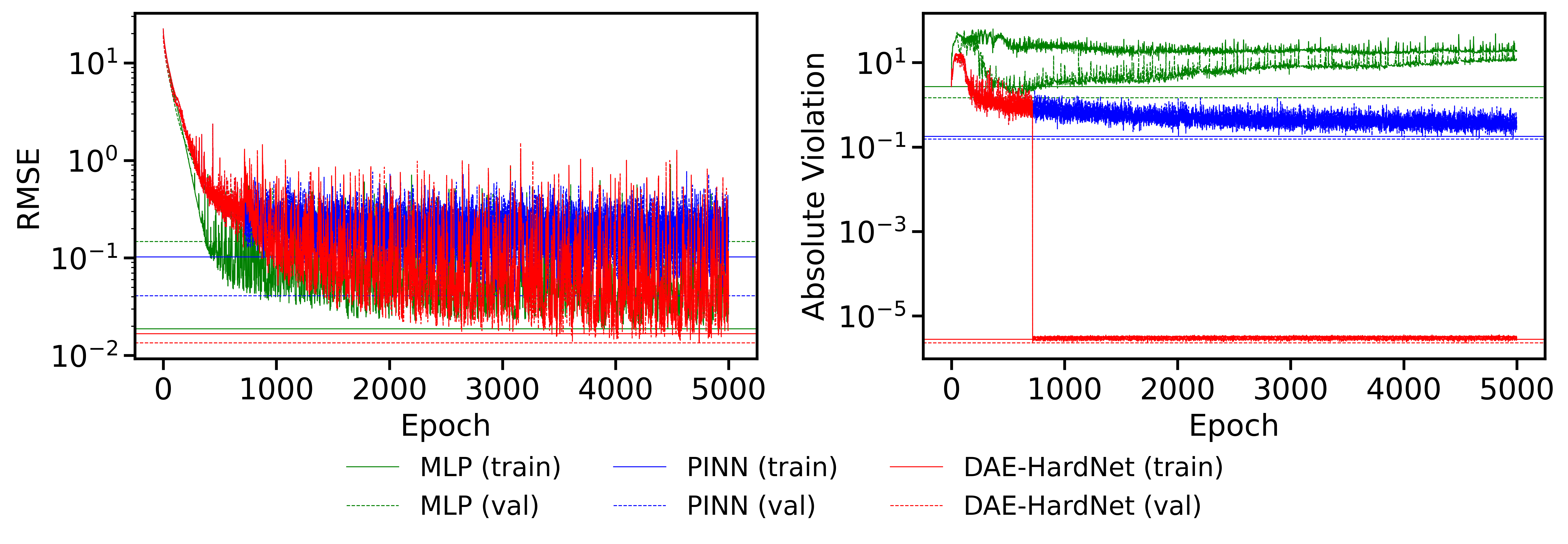}
    \caption{Learning curves for example of PDE in illustrative example 2. Left: MSE data loss; Right: absolute constraint violation on the projected gradients over 5000 epochs.}
    \label{fig:greenberg_pde_training}
\end{figure}

We use the solution given in Eq. \ref{eq:greeberg_sol1} to generate $2000$ data points over a uniform grid in $(x_1, x_2) \in [-1, 1] \times [-1, 1]$. Figure \ref{fig:greenberg_pde_prediction} shows the comparison of three models in terms of prediction. We observe that in spite of the predictions being very close to the ground truth values for all three models the absolute violation of the differential constraint in MLP and PINN is higher compared to DAE-HardNet. Figure \ref{fig:greenberg_pde_training} shows the RMSE data loss and absolute violation during training and testing for all three models. As mentioned in the training algorithm we first train the model based on PINN loss and then activate the projection layer dynamically when the total loss achieves the $\eta$ (here $0.05$) tolerance value. The sharp decrease in the absolute violation for DAE-HardNet in Figure \ref{fig:greenberg_pde_training} when the newton projection layer is activated.


The MSE data loss, MSE derivative loss and absolute violation achieved are reported in Table \ref{tab:pdemultisol}. Derivative loss of $9.29\times10^{-4}$ was achieved during training for the best data loss whereas the corresponding derivative loss during validation was found to be $2.17\times10^{-2}$. The constraint violation given is in terms of the projected derivative values which differ from the exact model derivatives. However, training the model on derivative loss minimizes the difference between the projected gradient value and the auto differentiation derivative values. Notably, the MSE data loss was achieved one order of magnitude lower than that of MLP and two orders of magnitude lower than that of PINN for both training and validation. This shows that our framework not only improves the derivative information, but also improves the data loss compared to MLP and PINN.

\begin{table}[htbp!]
    \centering
    \caption{Regression accuracy, constraint enforcement comparison and model configuration for the example of PDE with multiple solution in illustrative example 2.}
    \label{tab:pdemultisol}
    \begin{adjustbox}{width=\textwidth}
    \begin{tabular}{lccccccc}
    \toprule
    \multicolumn{1}{c}{\multirow{2}{*}{\textbf{Model}}} & \multicolumn{3}{c}{\textbf{Training}}                             & \multicolumn{3}{c}{\textbf{Validation}}                           & \multicolumn{1}{c}{\multirow{2}{*}{\textbf{Best Epoch}}} \\ 
    \cmidrule(lr){2-4}\cmidrule(lr){5-7}
    \multicolumn{1}{c}{}    & MSE (Data)                    & MSE (Derivative)          & Abs Violation                 & MSE (Data)                    & MSE (Derivative)          & Abs Violation                 & \multicolumn{1}{c}{}\\ \midrule
    MLP                     & \(1.03\!\times\!10^{-3}\)     & --                        & \(1.90\!\times\!10^{1}\)      & \(2.34\!\times\!10^{-2}\)     & --                        & \(1.13\!\times\!10^{1}\)      & 4990 \\
    PINN                    & \(4.31\!\times\!10^{-2}\)     & --                        & \(3.44\!\times\!10^{-1}\)     & \(2.66\!\times\!10^{-3}\)     & --                        & \(2.95\!\times\!10^{-1}\)     & 4390 \\
    DAEHN                   & \(7.49\!\times\!10^{-4}\)     & \(9.29\!\times\!10^{-4}\) & \(3.22\!\times\!10^{-6}\)     & \(2.17\!\times\!10^{-4}\)     & \(1.88\!\times\!10^{-2}\) & \(2.96\!\times\!10^{-6}\)     & 4820 \\ \midrule
    \multicolumn{8}{c}{\textbf{Model Configurations}}                                                                                                                                                                      \\ \midrule
    \multicolumn{2}{l}{\texttt{num\_epochs}}                & \multicolumn{2}{l}{5000}                                  & \multicolumn{2}{l}{\texttt{taylor\_order}}                & \multicolumn{2}{l}{2}                \\
    \multicolumn{2}{l}{\texttt{model\_depth}}               & \multicolumn{2}{l}{4}                                     & \multicolumn{2}{l}{\texttt{eta}}                          & \multicolumn{2}{l}{0.05}             \\
    \multicolumn{2}{l}{\texttt{hidden\_dim}}                & \multicolumn{2}{l}{32}                                    & \multicolumn{2}{l}{\texttt{newton\_step\_length}}         & \multicolumn{2}{l}{1.00}             \\
    \multicolumn{2}{l}{\texttt{lr}}                         & \multicolumn{2}{l}{0.001}                                 & \multicolumn{2}{l}{\texttt{max\_newton\_iter}}            & \multicolumn{2}{l}{10}               \\
    \multicolumn{2}{l}{\texttt{num\_points}}                & \multicolumn{2}{l}{2500 out of 2500}                      & \multicolumn{2}{l}{\texttt{noise\_std}}                   & \multicolumn{2}{l}{1.00}             \\
    \multicolumn{2}{l}{\texttt{pinn\_reg\_factor}}          & \multicolumn{2}{l}{1.00}                                  & \multicolumn{2}{l}{\texttt{noise\_mean}}                  & \multicolumn{2}{l}{0.00}             \\
    \multicolumn{2}{l}{\texttt{hardnet\_reg\_factor}}       & \multicolumn{2}{l}{1.00}                                  & \multicolumn{2}{l}{\texttt{noise\_scale}}                 & \multicolumn{2}{l}{0.01}             \\ 
    \multicolumn{2}{l}{\texttt{taylor\_offset}}              & \multicolumn{2}{l}{0.01}                                  & \multicolumn{2}{l}{\texttt{}}                             & \multicolumn{2}{l}{}                 \\ \bottomrule
    \end{tabular}
    \end{adjustbox}
\end{table}
\subsection{Example 6: 1--D Transient Heat Conduction}
\begin{figure}[htbp!]
    \centering
    \includegraphics[width=\linewidth]{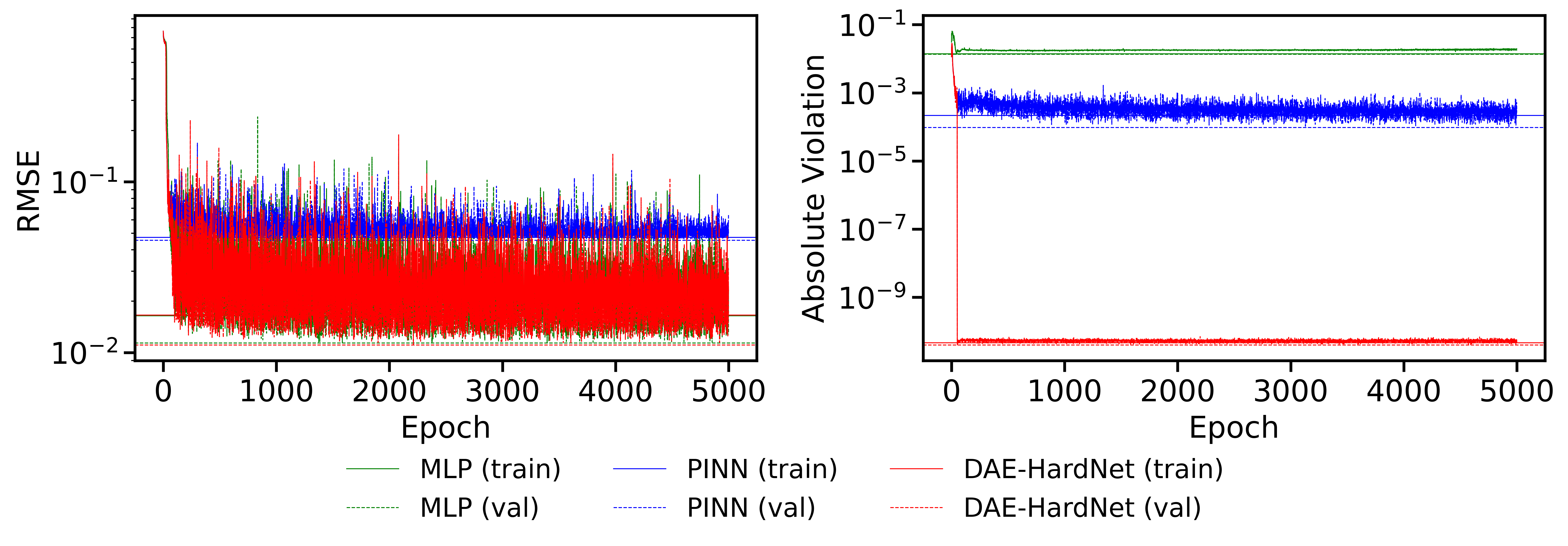}
    \caption{Learning curves for 1--D heat conduction problem. Left: MSE data loss; Right: absolute constraint violation on the projected gradients over 5000 epochs.}
    \label{fig:1d_heat_conduction_comparison}
\end{figure}

\begin{figure}[htbp!]
    \centering
    \includegraphics[width=\linewidth]{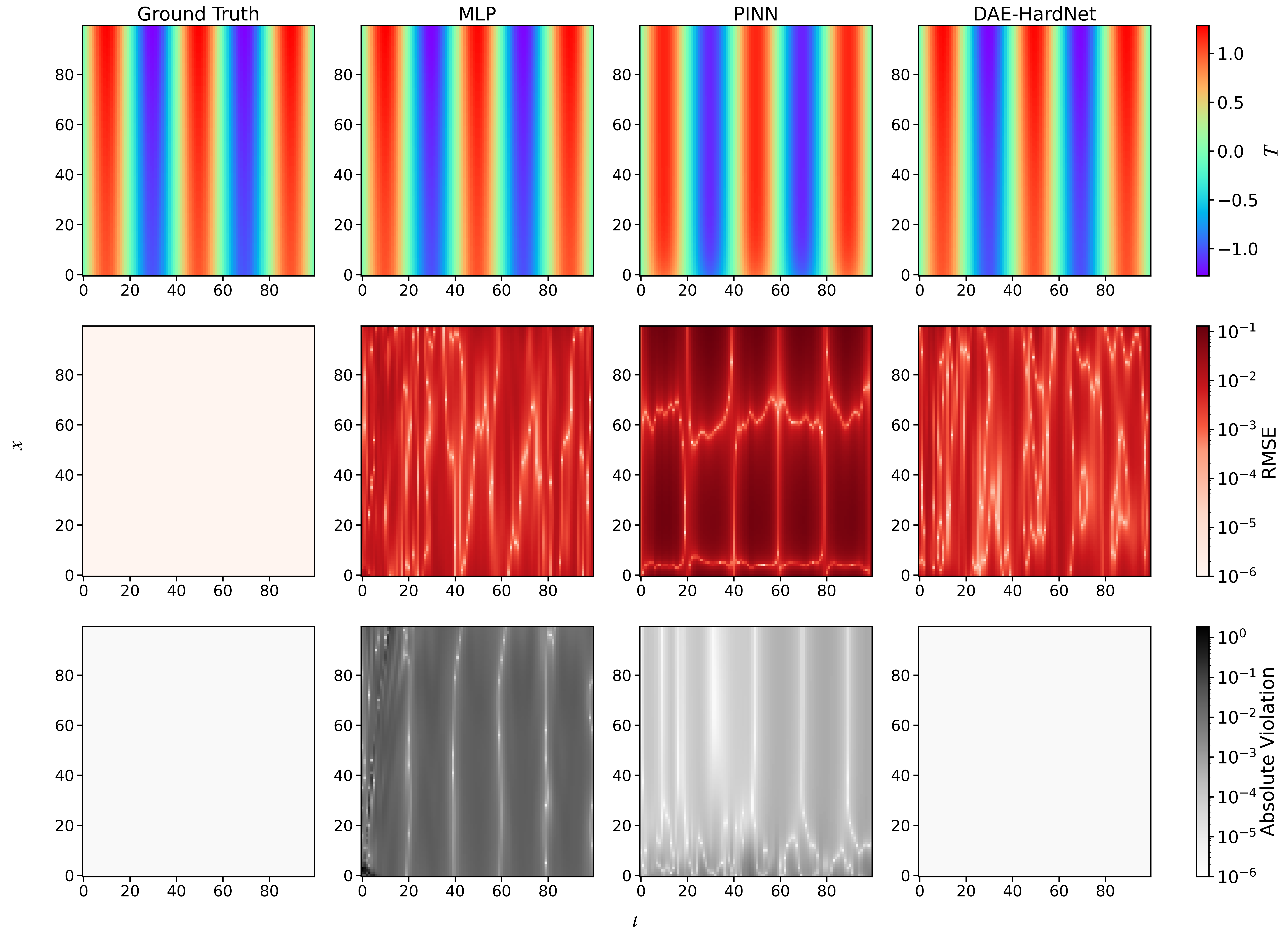}
    \caption{Model prediction comparison for 1--D heat conduction problem with ground truth. Top row: Model prediction heatmap; Middle row: Heatmap for the RMSE loss showing better performance of DAE-HardNet than MLP and PINN; Bottom row: Absolute violation heatmap with the model derivatives for MLP and PINN and with projected derivatives for DAE-HardNet.}
    \label{fig:1Dheatcomparison}
\end{figure}

We consider a $1$-D transient heat conduction problem of a thin rod of length $l$ which is not uniformly heated. The system is defined as,

\begin{ceqn}
    \begin{equation}
    \begin{aligned}
    \frac{\partial T}{\partial t} = \alpha\frac{\partial^2T}{\partial x^2}, &\quad \quad 0\leq x\leq l,t\geq0\\    
    \textrm{IC: } T(x,0) = \sin\left(\frac{n\pi x}{l}\right) &\quad \quad \textrm{BC: } T(0,t) = 0, T(l,t) = 0
    \end{aligned}
    \end{equation}
\end{ceqn}

where, $\alpha$ is the thermal diffusivity, and $n$ is the number of hot--cold zones along the length $l$. We considered the temporal domain $t\in[0,10]$ and spatial domain $x\in[0,5]$ to generate the data with $\alpha = 1$ and $n=5$. We considered 5000 points out of 10000 data points to train the models for 5000 epochs. Figure \ref{fig:1d_heat_conduction_comparison} provides the learning curves of data loss and absolute constraint violation on projected gradient for the heat conduction problem. Figure \ref{fig:1Dheatcomparison} provides the visual representation of the predicted temperature heatmap, the heatmap of data loss and the constraint violation for all models. For MLP and PINN the derivatives are the model derivatives through auto differentiation and for DAE-HardNet the constraint violation is in terms of projected derivative values.

\begin{table}[htbp!]
    \centering
    \caption{Regression accuracy, constraint enforcement comparison and model configuration for 1--D transient heat conduction system.}
    \label{tab:1DHeatConduction}
    \begin{adjustbox}{width=\textwidth}
    \begin{tabular}{lccrrrrrr}
    \toprule
    \multicolumn{1}{c}{\multirow{2}{*}{\textbf{Model}}} & \multicolumn{3}{c}{\textbf{Training}}                             & \multicolumn{3}{c}{\textbf{Validation}}                           & \multicolumn{1}{c}{\multirow{2}{*}{\textbf{Best Epoch}}} \\ 
    \cmidrule(lr){2-4}\cmidrule(lr){5-7}
    \multicolumn{1}{c}{}    & MSE (Data)                    & MSE (Derivative)          & Abs Violation                 & MSE (Data)                    & MSE (Derivative)          & Abs Violation                 & \multicolumn{1}{c}{}\\ \midrule
    MLP                     & \(4.26\!\times\!10^{-4}\)     & --                        & \(1.87\!\times\!10^{-2}\)     & \(1.40\!\times\!10^{-4}\)     & --                        & \(1.75\!\times\!10^{-2}\)     & 4240 \\
    PINN                    & \(2.36\!\times\!10^{-3}\)     & --                        & \(3.74\!\times\!10^{-4}\)     & \(2.09\!\times\!10^{-3}\)     & --                        & \(2.33\!\times\!10^{-4}\)     & 3650 \\
    DAEHN                   & \(4.77\!\times\!10^{-4}\)     & \(2.95\!\times\!10^{-5}\) & \(5.30\!\times\!10^{-11}\)    & \(1.37\!\times\!10^{-4}\)     & \(2.97\!\times\!10^{-5}\) & \(4.84\!\times\!10^{-11}\)    & 1560 \\ \midrule
    \multicolumn{8}{c}{\textbf{Model Configurations}}                                                                                                                                                                      \\ \midrule
    \multicolumn{2}{l}{\texttt{num\_epochs}}                & \multicolumn{2}{l}{5000}                                  & \multicolumn{2}{l}{\texttt{taylor\_order}}                & \multicolumn{2}{l}{2}                \\
    \multicolumn{2}{l}{\texttt{model\_depth}}               & \multicolumn{2}{l}{4}                                     & \multicolumn{2}{l}{\texttt{eta}}                          & \multicolumn{2}{l}{0.005}            \\
    \multicolumn{2}{l}{\texttt{hidden\_dim}}                & \multicolumn{2}{l}{32}                                    & \multicolumn{2}{l}{\texttt{newton\_step\_length}}         & \multicolumn{2}{l}{1.00}             \\
    \multicolumn{2}{l}{\texttt{lr}}                         & \multicolumn{2}{l}{0.001}                                 & \multicolumn{2}{l}{\texttt{max\_newton\_iter}}            & \multicolumn{2}{l}{5}               \\
    \multicolumn{2}{l}{\texttt{num\_points}}                & \multicolumn{2}{l}{5000 out of 10000}                     & \multicolumn{2}{l}{\texttt{noise\_std}}                   & \multicolumn{2}{l}{1.00}             \\
    \multicolumn{2}{l}{\texttt{pinn\_reg\_factor}}          & \multicolumn{2}{l}{1.00}                                  & \multicolumn{2}{l}{\texttt{noise\_mean}}                  & \multicolumn{2}{l}{0.00}             \\
    \multicolumn{2}{l}{\texttt{hardnet\_reg\_factor}}       & \multicolumn{2}{l}{1.00}                                  & \multicolumn{2}{l}{\texttt{noise\_scale}}                 & \multicolumn{2}{l}{0.01}             \\
    \multicolumn{2}{l}{\texttt{taylor\_offset}}             & \multicolumn{2}{l}{0.1}                                   & \multicolumn{2}{l}{\texttt{}}                             & \multicolumn{2}{l}{}                 \\ \bottomrule
    \end{tabular}
    \end{adjustbox}
\end{table}

The evaluation metrics and model configurations are provided in Table \ref{tab:1DHeatConduction}. Derivative loss around $3\times10^{-5}$ were achieved for both training and testing. MSE data loss was achieved close to MLP model. This shows how DAE-HardNet is able to achieve the best traits of MLP and PINN model with data loss as close as MLP or even better. It can also reduce the physics loss by learning the derivatives that satisfy the equations.

\subsection{Sensitivity Analysis}

We performed sensitivity analysis for all models on the number of data points used for training, the noise present in the data and the offset of the Taylor projection mentioned in section \ref{sec:daehardnet-framework}. For the analysis on the scale of noise we generated synthetic noise from gaussian distribution of mean $0$ and standard deviation $1$ multiplied by the scale of the noise. Table \ref{tab:sensitivity_analysis} shows the results of the sensitivity analysis. As the number of data points used for training decreases we observed that the data loss of all three models increased. Compared to MLP and PINN, DAE-HardNet achieved lower or comparative data loss for both training and validation set. Contrastively for validation set with small number of data points we observed increase in the derivative loss. This could be either due to the backbone not being able to generalize with smaller number of data points or due to the inefficient projection considering small number of data points. With increasing noise in the data we observed DAE-HardNet achieves lowest data loss among all the models since it satisfied the constraints of the system. Additionally, with decreasing Taylor offset the data loss of DAE-HardNet decreased which was most likely due to the linearization around smaller neighborhood for the Taylor projection.

\textit{\textbf{Remark 4}} Considering a very small Taylor offset can create numerical instability in the projection layer due to the possibility of the jacobian of KKT system being singular. We recommend to consider a offset value between $0.001$ and $0.1$.
\begin{table}[htbp!]
\centering
\caption{Regression accuracy and constraint enforcement analysis under varying model parameters for illustrative example 1.}
\label{tab:sensitivity_analysis}
\begin{adjustbox}{width=\textwidth}
\begin{tabular}{lclcccccc}
\toprule
\textbf{Parameter} & \textbf{Setting} & \textbf{Model} 
& \multicolumn{3}{c}{\textbf{Training}} 
& \multicolumn{3}{c}{\textbf{Validation}} \\
\cmidrule(lr){4-6}\cmidrule(lr){7-9}
& & & MSE (Data) & MSE (Derivative) & Abs Violation & MSE (Data) & MSE (Derivative) & Abs Violation \\
\midrule

\multirow{9}{*}{\texttt{num\_points}} 
    & \multirow{3}{*}{1000} & MLP       & \(8.55\!\times\!10^{-5}\) &             --            & \(3.59\!\times\!10^{0}\) & \(2.95\!\times\!10^{-5}\)  &             --            &\(3.27\!\times\!10^{0}\) \\
    &                       & PINN      & \(1.43\!\times\!10^{-3}\) &             --            & \(1.44\!\times\!10^{-1}\) & \(4.06\!\times\!10^{-4}\) &             --            &\(7.95\!\times\!10^{-2}\) \\
    &                       & DAEHN     & \(7.14\!\times\!10^{-6}\) & \(2.53\!\times\!10^{-4}\) & \(9.50\!\times\!10^{-7}\) & \(4.57\!\times\!10^{-6}\) & \(2.79\!\times\!10^{-4}\) &\(8.96\!\times\!10^{-7}\) \\
    \cmidrule(lr){2-9}

    & \multirow{3}{*}{500}  & MLP       & \(1.21\!\times\!10^{-4}\) &             --            & \(4.70\!\times\!10^{0}\) & \(1.17\!\times\!10^{-4}\)  &             --            & \(3.84\!\times\!10^{0}\) \\
    &                       & PINN      & \(2.75\!\times\!10^{-3}\) &             --            & \(1.66\!\times\!10^{-1}\) & \(7.3\!\times\!10^{-4}\)  &             --            & \(2.17\!\times\!10^{-1}\) \\
    &                       & DAEHN     & \(3.35\!\times\!10^{-5}\) & \(3.87\!\times\!10^{-4}\) & \(9.63\!\times\!10^{-7}\) & \(9.22\!\times\!10^{-6}\) & \(3.47\!\times\!10^{-4}\) & \(9.54\!\times\!10^{-7}\) \\
    \cmidrule(lr){2-9}

    & \multirow{3}{*}{100}  & MLP           & \(3.98\!\times\!10^{-3}\) &             --            & \(9.14\!\times\!10^{0}\) & \(1.20\!\times\!10^{-3}\) &             --            & \(1.19\!\times\!10^{1}\) \\
    &                       & PINN          & \(7.46\!\times\!10^{-2}\) &              --           & \(5.82\!\times\!10^{-1}\) & \(7.70\!\times\!10^{-2}\) &             --            & \(1.16\!\times\!10^{0}\) \\
    &                       & DAEHN         & \(4.21\!\times\!10^{-3}\) & \(3.67\!\times\!10^{-3}\) & \(1.17\!\times\!10^{-6}\) & \(1.10\!\times\!10^{-3}\) & \(6.10\!\times\!10^{0}\) & \(8.46\!\times\!10^{-7}\) \\
\midrule

\multirow{9}{*}{\texttt{noise\_scale}} 
    & \multirow{3}{*}{0.001}    & MLP       & \(1.71\!\times\!10^{-5}\) &             --            & \(2.46\!\times\!10^{0}\) & \(1.10\!\times\!10^{-5}\)  &             --            & \(2.10\!\times\!10^{0}\) \\
    &                           & PINN      & \(2.25\!\times\!10^{-3}\) &             --            & \(2.03\!\times\!10^{-1}\) & \(4.24\!\times\!10^{-4}\) &             --            & \(1.01\!\times\!10^{-1}\) \\
    &                           & DAEHN     & \(5.58\!\times\!10^{-6}\) & \(2.22\!\times\!10^{-4}\) & \(9.88\!\times\!10^{-7}\) & \(2.09\!\times\!10^{-6}\) & \(2.33\!\times\!10^{-4}\) & \(1.02\!\times\!10^{-6}\) \\
    \cmidrule(lr){2-9}

    & \multirow{3}{*}{0.01} & MLP       & \(3.46\!\times\!10^{-4}\) &             --            & \(2.59\!\times\!10^{0}\)  & \(1.27\!\times\!10^{-4}\) &             --            & \(2.10\!\times\!10^{0}\) \\
    &                       & PINN      & \(1.74\!\times\!10^{-3}\) &             --            & \(1.59\!\times\!10^{-1}\) & \(4.06\!\times\!10^{-4}\) &             --            & \(7.95\!\times\!10^{-2}\) \\
    &                       & DAEHN     & \(2.22\!\times\!10^{-4}\) & \(3.74\!\times\!10^{-4}\) & \(9.45\!\times\!10^{-7}\) & \(9.71\!\times\!10^{-5}\) & \(3.11\!\times\!10^{-4}\) & \(9.03\!\times\!10^{-7}\) \\
    \cmidrule(lr){2-9}

    & \multirow{3}{*}{0.1}  & MLP       & \(2.44\!\times\!10^{-2}\) &             --            & \(4.87\!\times\!10^{0}\) & \(9.58\!\times\!10^{-3}\) &             --            & \(4.46\!\times\!10^{0}\) \\
    &                       & PINN      & \(2.19\!\times\!10^{-2}\) &             --            & \(2.46\!\times\!10^{-1}\) & \(9.28\!\times\!10^{-3}\) &             --            & \(1.40\!\times\!10^{-1}\) \\
    &                       & DAEHN     & \(2.11\!\times\!10^{-2}\) & \(2.36\!\times\!10^{-3}\) & \(9.41\!\times\!10^{-7}\) & \(9.01\!\times\!10^{-3}\) & \(1.65\!\times\!10^{-3}\) & \(9.31\!\times\!10^{-7}\) \\
\midrule

\multirow{9}{*}{\texttt{taylor\_offset}} 
    & \multirow{3}{*}{0.1}  & MLP       & \(2.46\!\times\!10^{-5}\) &             --            & \(2.96\!\times\!10^{0}\) & \(1.88\!\times\!10^{-5}\) &             --            & \(2.69\!\times\!10^{0}\) \\
    &                       & PINN      & \(1.41\!\times\!10^{-3}\) &             --            & \(1.27\!\times\!10^{-1}\) & \(5.02\!\times\!10^{-4}\) &             --            & \(1.32\!\times\!10^{-1}\) \\
    &                       & DAEHN     & \(7.93\!\times\!10^{-6}\) & \(2.59\!\times\!10^{-4}\) & \(9.72\!\times\!10^{-7}\) & \(4.29\!\times\!10^{-6}\) & \(2.98\!\times\!10^{-4}\) & \(9.16\!\times\!10^{-7}\) \\
    \cmidrule(lr){2-9}

    & \multirow{3}{*}{0.01} & MLP       & \(2.46\!\times\!10^{-5}\) &             --            & \(2.96\!\times\!10^{0}\)  & \(1.88\!\times\!10^{-5}\) &             --            & \(2.69\!\times\!10^{0}\) \\
    &                       & PINN      & \(1.41\!\times\!10^{-3}\) &             --            & \(1.27\!\times\!10^{-1}\) & \(5.02\!\times\!10^{-4}\) &             --            & \(1.32\!\times\!10^{-1}\) \\
    &                       & DAEHN     & \(3.25\!\times\!10^{-5}\) & \(2.53\!\times\!10^{-4}\) & \(1.08\!\times\!10^{-6}\) & \(5.69\!\times\!10^{-6}\) & \(3.74\!\times\!10^{-4}\) & \(1.06\!\times\!10^{-6}\) \\
    \cmidrule(lr){2-9}

    & \multirow{3}{*}{0.001}    & MLP       & \(2.46\!\times\!10^{-5}\) &             --            & \(2.96\!\times\!10^{0}\)  & \(1.88\!\times\!10^{-5}\) &             --            & \(2.70\!\times\!10^{0}\) \\
    &                           & PINN      & \(1.41\!\times\!10^{-3}\) &             --            & \(1.27\!\times\!10^{-1}\) & \(5.02\!\times\!10^{-4}\) &             --            & \(1.32\!\times\!10^{-1}\) \\
    &                           & DAEHN     & \(2.35\!\times\!10^{-6}\) & \(4.34\!\times\!10^{-5}\) & \(9.83\!\times\!10^{-7}\) & \(9.04\!\times\!10^{-7}\) &\(3.04\!\times\!10^{-5}\) & \(8.66\!\times\!10^{-7}\) \\
\bottomrule
\end{tabular}
\end{adjustbox}
\end{table}

\subsection{Computational Time Analysis}

\begin{figure}[htbp!]
    \centering
    \includegraphics[width=\linewidth]{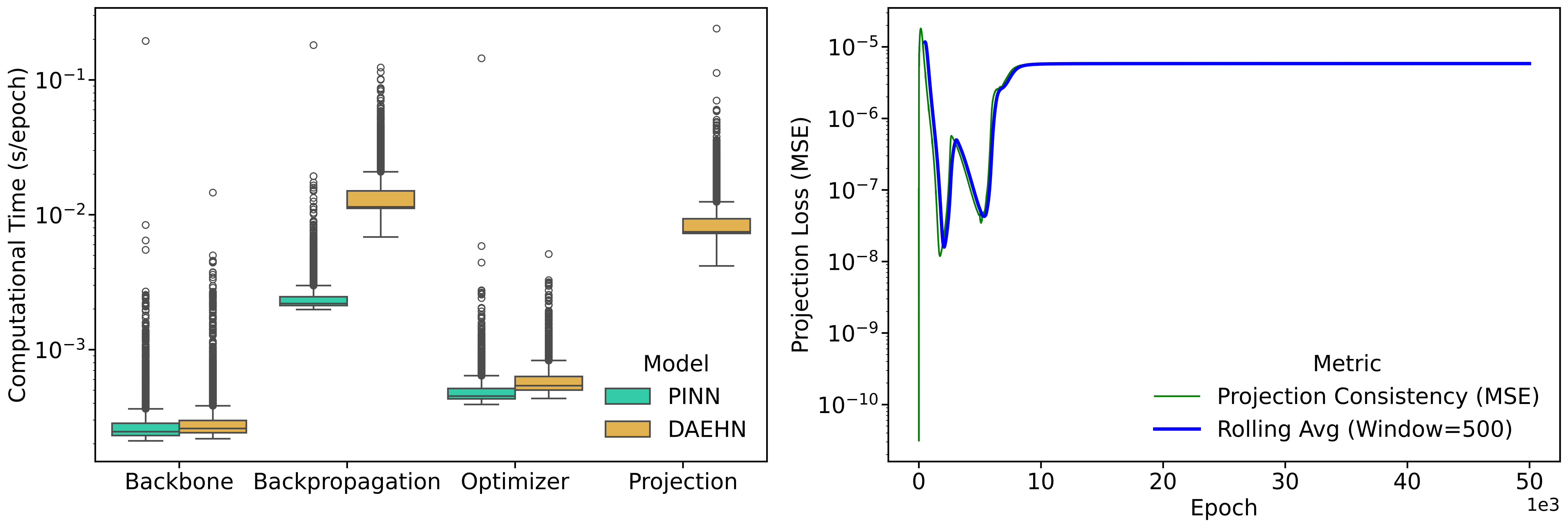}
    \caption{Computational Analysis for DAE-HardNet. Left: Computational time for different parts of the framework for Lotka-Volterra System. Right: Difference between the model outputs before and after projection layer. Low projection loss corresponds to improvement in the backbone prediction towards constrained outputs.}
    \label{fig:computational_comparison}
\end{figure}

We conducted computational analysis of all three models for ``Lotka-Volterra" case study. Figure \ref{fig:computational_comparison}(left) shows the average time required in seconds per epoch of computation for different segments of the framework. During training, the MLP and PINN architectures did not require the projection layer, and their gradient updates rely solely on standard auto-differentiation of the backbone network. In contrast, DAE-HardNet required additional computations for both the projection layer and the associated model gradients at each training step. However, during inference, only the backbone and projection components were evaluated for DAE-HardNet, whereas for MLP and PINN, only the backbone network is active. 


DAE-HardNet is generally computationally more expensive compared to typical MLP and PINNs. This is primarily due to the projection layer being the most time-consuming steps of the entire framework. Also, the time required for the projection layer is found to be closely proportional to the problem dimensionality. So, to reduce the computational complexity of the overall framework, we would like minimize the difference between the unconstrained outputs of the backbone NN and constrained outputs from the projection layer. Figure \ref{fig:computational_comparison} shows the difference between unconstrained and projected constrained variables over epochs. Since we used minimum distance projection of the unconstrained outputs on the constraint space, the difference between the unconstrained outputs and constrained outputs were on the order of $10^{-6}$. Considering such low order of difference we could remove the projection layer during inference. Using this approach, the computational complexity of MLP, PINN and DAE-HardNet is expected to be similar during inference.


\section{Conclusions}
\label{sec:conclusions}
Modeling differential-algebraic (DAE) systems are comparatively more challenging than modeling a system of only differential equations \cite{petzold1982differential} and are substantially harder than modeling an algebraic system. In this work, we presented a physics constrained neural network architecture that is able to strictly enforce DAE constraints along with hard nonlinear/linear algebraic equalities and inequalities in both inputs and outputs through a differentiable projection layer. During projection, the terms with differential operators of state variables are also considered as new individual variables that are connected to the original state variables through \textit{multiple-point neighborhood approach} of Taylor expansion. The projection layer involved solving a squared system of equations corresponding to the KKT conditions of a distance minimization problem. In doing so, we addressed the limitations of traditional PINNs that rely on regularization parameter and lacks guarantee of first principle satisfaction. Through the projection layer we enforced hard constraint satisfaction on the projected derivatives for general nonlinear systems. We then used the loss function to minimize the loss for data and the model derivative with the projected derivative. From the case studies we observed that the model is able to achieve data loss on the same order or lower as MLP with constraint violation close to machine precision on the projected derivatives. Additionally, minimizing the derivative loss improved the derivative information of auto-differentiation and created synergistic effect on the backbone parameters. DAE-HardNet was also able to capture accurate derivative information, whereas MLP and PINN struggled to do so. Sensitivity analysis suggested that considering smaller Taylor offset might lead to even more improved results due to high fidelity linearization of the Taylor projection. The training of DAE-HardNet was comparatively expensive due to the projection layer. However during inference, the projection layer could be removed if the differences between the unconstrained outputs of the neural network backbone and the constrained outputs of projection layer were lower than the prediction tolerance value. With this consideration, the projection layer and the training scheme improved the prediction capability of the backbone compared to MLP and PINN and only the backbone could be used during inference.

DAE-HardNet shows significant promise to enable scientific machine learning and incorporation of domain knowledge in data-driven surrogate models for a wider arrays of scientific problems  characterized by DAE-based governing equations. ANNs have been traditionally used to approximate nonlinear functions and have shown exceptional performance across a variety of domains like computer vision and natural language processing. However, for engineering domain problems, they often suffer from lack of interpretability. To that end, DAE-HardNet can be extended for solving large-scale DAE systems for inverse modeling, large-scale process synthesis, dynamic optimization, real-time control, molecular and process design, and many more.

\section*{Acknowledgments}
The authors gratefully acknowledge partial funding support from the NSF CAREER award CBET-1943479 and the EPA Project Grant 84097201.

\printbibliography



\end{document}